\documentclass[10pt,twocolumn,letterpaper]{article}

\usepackage{iccv}
\usepackage{times}
\usepackage{epsfig}
\usepackage{graphicx}
\usepackage{amssymb}
\usepackage{amsmath,amsfonts,amsthm}
\usepackage[accsupp]{axessibility} 
\usepackage{booktabs}
\usepackage{algorithmicx}
\usepackage{algorithm}
\usepackage{xcolor}
\usepackage{color}
\usepackage{etoolbox}
\usepackage{subcaption}
\usepackage{multirow}
\usepackage{textcomp}
\usepackage[noend]{algpseudocode}
\usepackage[normalem]{ulem}
\usepackage{bbding}
\usepackage{svg}
\usepackage{colortbl}  

\usepackage{array}

\usepackage[breaklinks=true,bookmarks=false,colorlinks=true,linkcolor=red,citecolor=green,urlcolor=cyan]{hyperref}

\usepackage[capitalize]{cleveref}
\crefname{section}{Sec.}{Secs.}
\Crefname{section}{Section}{Sections}
\Crefname{table}{Table}{Tables}
\crefname{table}{Tab.}{Tabs.}

\iccvfinalcopy 

\def\iccvPaperID{9416} 

\ificcvfinal\fi

\begin{document}

\title{QD-BEV : Quantization-aware View-guided Distillation for Multi-view \\3D Object Detection}

\author{
Yifan Zhang\textsuperscript{1*},\and
Zhen Dong\textsuperscript{2*},\and
Huanrui Yang\textsuperscript{2},\and
Ming Lu\textsuperscript{3},\and
Cheng-Ching Tseng\textsuperscript{3},\and
Yuan Du\textsuperscript{1},\and
Kurt Keutzer\textsuperscript{2},\and
Li Du\textsuperscript{1†},\and
Shanghang Zhang\textsuperscript{3†}
\\
\small\textsuperscript{1}Nanjing University,
\small\textsuperscript{2}University of California, Berkeley,\\
\small\textsuperscript{3}National Key Laboratory for Multimedia Information Processing, School of Computer Science, Peking University,\\
}
\maketitle

\ificcvfinal\thispagestyle{empty}\fi


\begin{figure}[b]
\rule{0.25\textwidth}{0.4pt}
\vspace{2pt}
\small
\begin{tabular}{@{}l}
* Equal contribution :\\
zhang\_yifan@smail.nju.edu.cn, zhendong@berkeley.edu \\
† Corresponding authors :\\
\small ldu@nju.edu.cn, shanghang@pku.edu.cn
\end{tabular}
\end{figure}

\begin{abstract}

Multi-view 3D detection based on BEV (bird-eye-view) has recently achieved significant improvements.
However, the huge memory consumption of state-of-the-art models makes it hard to deploy them on vehicles, and the non-trivial latency will affect the real-time perception of streaming applications.
Despite the wide application of quantization to lighten models, we show in our paper that directly applying quantization in BEV tasks will 1) make the training unstable, and 2) lead to intolerable performance degradation.
To solve these issues, our method QD-BEV enables a novel view-guided distillation (VGD) objective, which can stabilize the quantization-aware training (QAT) while enhancing the model performance by leveraging both image features and BEV features.
Our experiments show that QD-BEV achieves similar or even better accuracy than previous methods with significant efficiency gains. On the nuScenes datasets, the 4-bit weight and 6-bit activation quantized QD-BEV-Tiny model achieves 37.2\% NDS with only 15.8 MB model size, outperforming BevFormer-Tiny by 1.8\% with an 8$\times$ model compression. On the Small and Base variants, QD-BEV models also perform superbly and achieve 47.9\% NDS (28.2 MB) and 50.9\% NDS (32.9 MB), respectively.
\end{abstract}

\section{Introduction}
\label{sec:intro}
\renewcommand{\dblfloatpagefraction}{.9}

\begin{figure}[t]
  \centering
  \begin{subfigure}{0.49\linewidth}
    \includegraphics[width=4.3cm, trim={0 0 0 0}, clip]{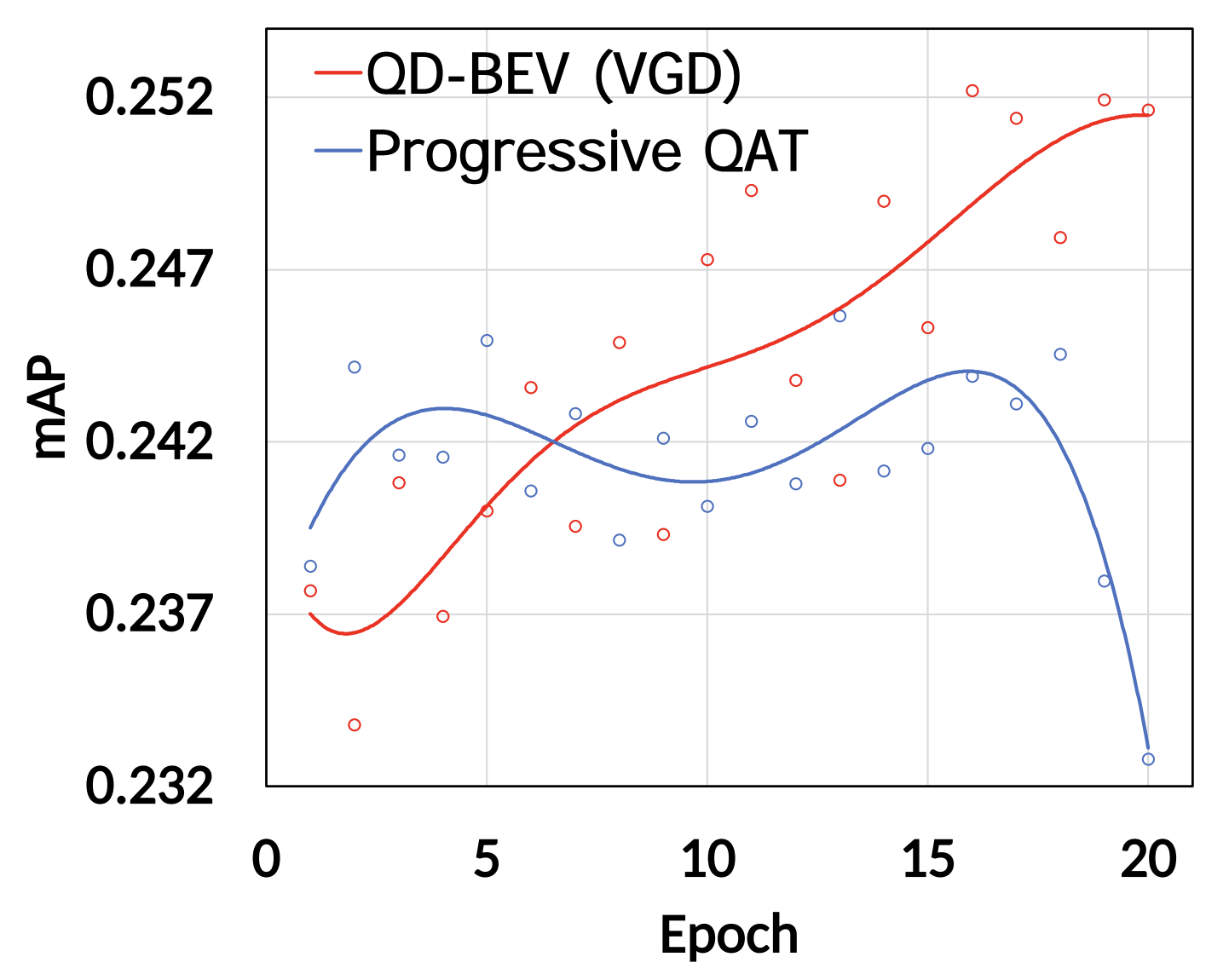}
    \caption{mAP curve.}
    \label{fig:tiny_map}
  \end{subfigure}
  \hfill
  \begin{subfigure}{0.49\linewidth}
    \includegraphics[width=4.3cm, trim={0 0 0 0}, clip]{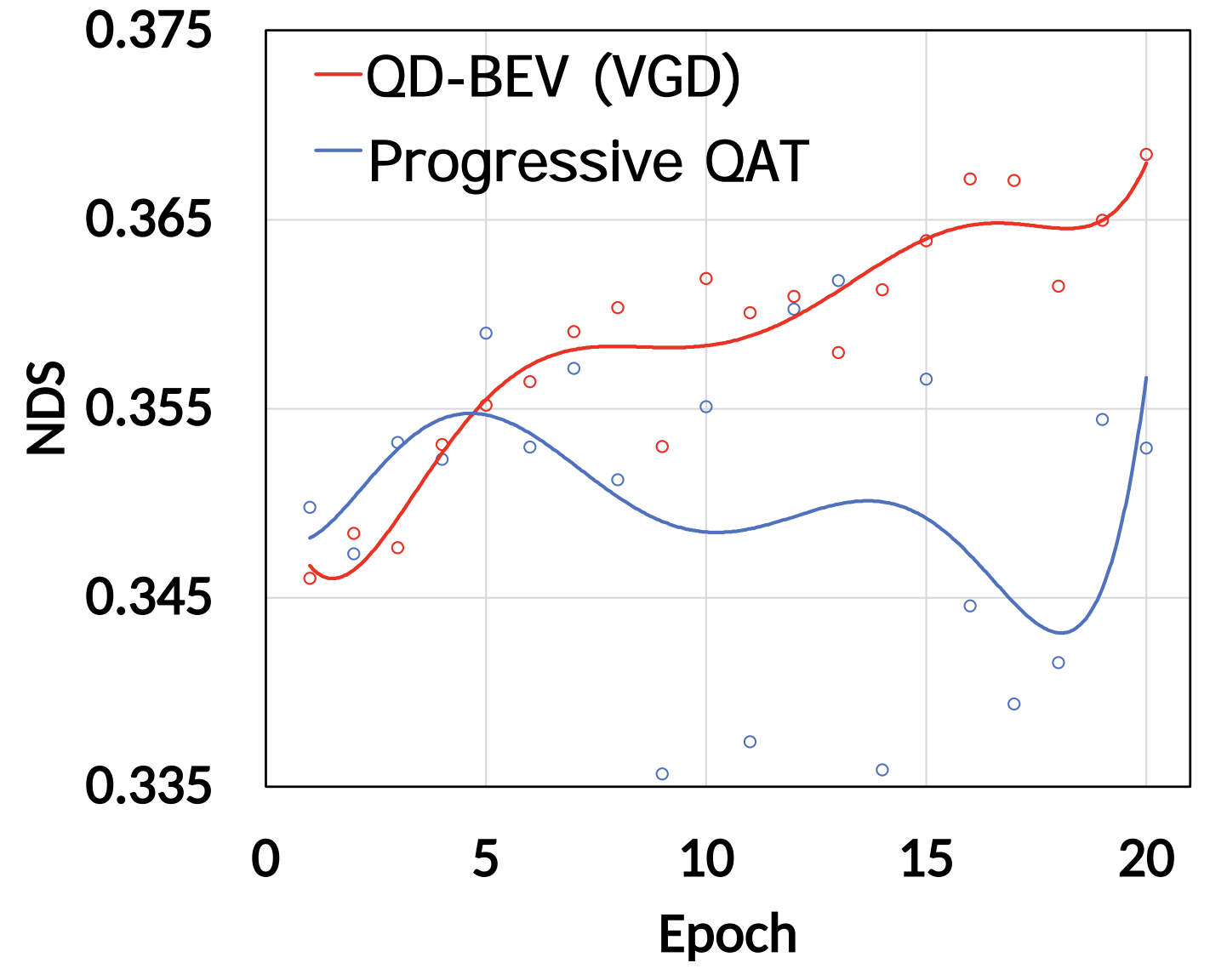}
    \caption{NDS curve.}
    \label{fig:tiny_nds}
  \end{subfigure}
    \centering
  \caption{Training curves of QD-BEV (with VGD) versus progressive QAT on W4A6 quantization of BEVFormer-Tiny. Note that standard QAT works even worse compared to progressive QAT and it falls out of the targeted accuracy range in the figures.}
  \label{fig:vgd curve}
\end{figure}

\begin{figure*}[htp]
\centering
      \includegraphics[width=1\linewidth]{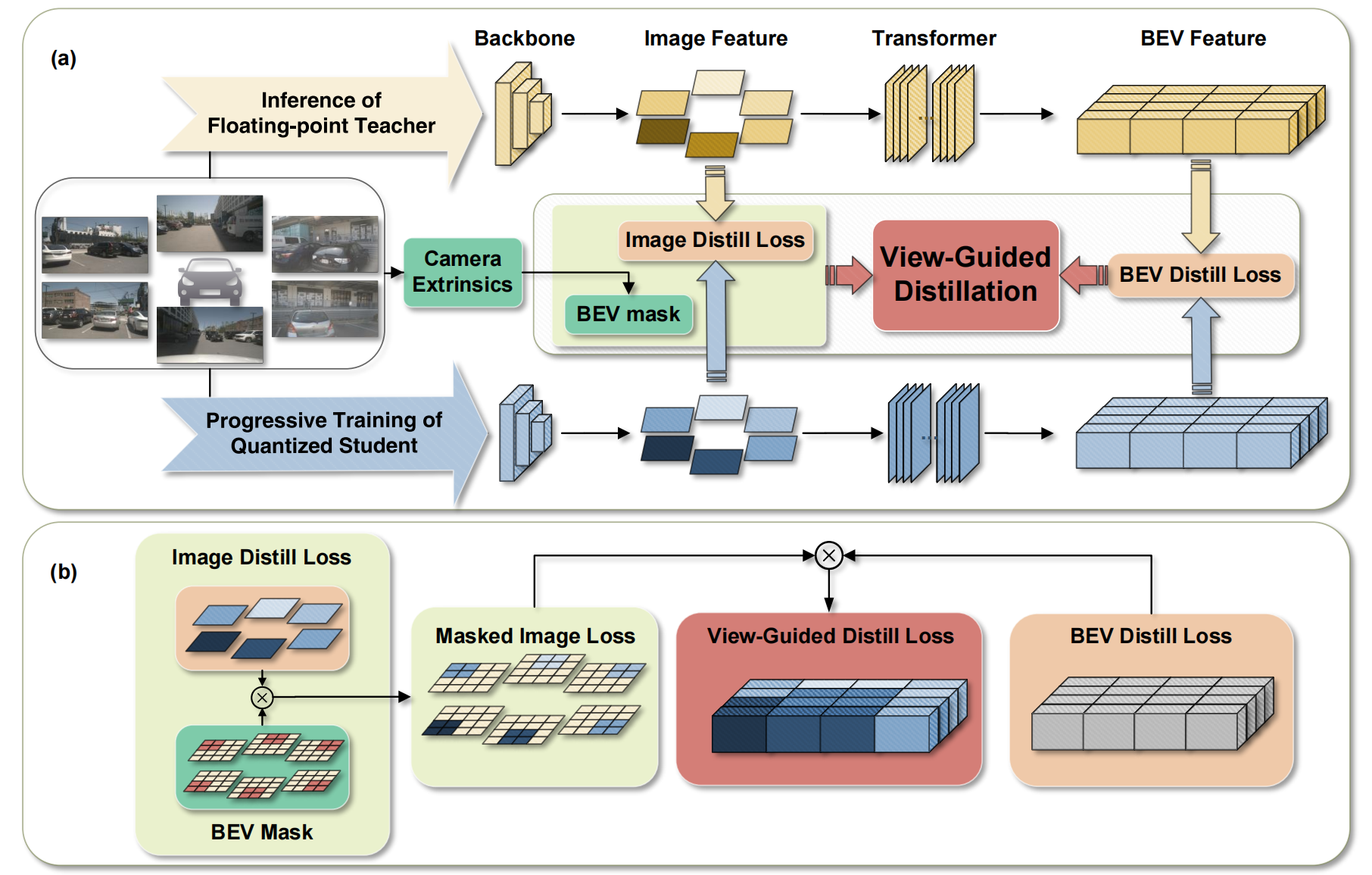}
   \caption{\textbf{Illustration of QD-BEV.} (a) In our pipeline, multi-camera images are input into the floating-point teacher network and the quantized student network in order to compute the KL divergence in an element-wise manner. The KL divergence is used as distillation loss in the image feature and the BEV feature, respectively. Then we conduct view-guided distillation using the BEV mask obtained from the external parameters of the camera.
   Please refer to Sec.\ref{subsec:overview of qd-bev} for more details.
   (b) The lower flow chart shows the computation process of the view-guided distillation loss.
   Specific details are in Sec. \ref{subsec:vgd}.}
   \label{fig:intro}
\end{figure*}
















Given its potential in enabling autopilot, multi-view 3D detection based on BEV (bird-eye-view)
has become an important research direction for autonomous driving. 
Based on input sensors, previous work can be divided into LiDAR-based methods \cite{lang2019pointpillars,zhou2018voxelnet} and camera-only methods \cite{li2022bevformer,wang2022detr3d,huang2021bevdet,huang2022bevdet4d,liu2022petr,liu2022petrv2}.
Compared to the LiDAR-based methods, camera-only methods have the merits of lower deployment cost, closer to human eyes, and easier access to visual information in the driving environment.
However, even if using the camera-only methods, the computational and memory costs of running state-of-the-art BEV models are still formidable, making it difficult to deploy them onto vehicles. For example, 
BEVFormer-Base has a 540 ms inference latency (corresponds to 1.85 fps) on one NVIDIA V100 GPU, which is infeasible for real-time applications that generally require 30 fps.
Since a non-trivial latency will harm the streaming perception, it is particularly crucial to explore and devise lightweight models for camera-only 3D object detection based on BEV.

Quantization~\cite{jacob2018quantization, gholami2021survey, zhang_2018_lqnets, dong2019hawq} can reduce the bitwidth used to represent weights and activations in deep neural networks, which can greatly save the model size and computational costs while improving the speed of model reasoning.
However, directly applying quantization would lead to significant performance degradation. 
Compared to image classification and 2D object detection tasks where the standard quantization methods shine, multi-camera 3D detection tasks are much more complicated and difficult due to the existence of multiple views and information from multiple dimensions (for example, the temporal information and spatial information used in BEVFormer~\cite{li2022bevformer}). Consequently, the structure of BEV networks tends to become more complex, with a deeper convolutional neural network backbone to extract image information from multiple views, and with transformers to encode and decode the features of the BEV domain. The presence of different neural architectures, multiple objectives, and knowledge from different modalities greatly challenges the standard quantization methods, decreasing their stability and accuracy, and even making the whole training process diverge. In Figure~\ref{fig:vgd curve}, we show the training curves when applying W4A6 quantization on a BEVFormer-Tiny model. As can be seen, the performance of quantization-aware training (QAT) fluctuates significantly in different epochs, while the performance of our proposed method QD-BEV shows a stable rising trend. We perform more experiments to validate the effectiveness of QD-BEV in Sec~\ref{sec:ablation_study}.





To solve the problems of standard QAT, in this work, we first conduct systematic experiments and analyses on quantizing BEV networks. Then we devise a quantization-aware view-guided distillation method (referred to as QD-BEV) that can decently solve the stability issue while improving the final performance of compact BEV models.
Our proposed view-guided distillation (VGD) can better leverage information from both the image and the BEV domains for multi-view 3D object detection. This can significantly outperform previous distillation methods that cannot jointly handle the different types of losses in BEV networks.
Specifically, as shown in Figure~\ref{fig:intro}, we first take the FP (floating-point) model as the teacher model and the quantized model as the student model, then we calculate the KL divergence on the image feature and the BEV feature, respectively. 
Finally, we leverage the mapping relationship and realize VGD by organically combining the image feature and the BEV feature through the camera's external parameters. Note that in QD-BEV, neither additional training data nor larger powerful teacher networks are used to tune the accuracy, but QD-BEV models are still able to outperform previous baselines while having significantly smaller model sizes and computational requirements.
Our contributions are as follows:
\begin{itemize}
    \item We conduct systematic experiments on quantizing BEV models, unveiling major issues hampering standard quantization-aware training methods on BEV. 
    \item We specially design view-guided distillation (VGD) for BEV models, which jointly leverages both image domain and BEV domain information. VGD boosts the final performance while solving the stability issue of standard QAT. 
    \item Our W4A6 quantized QD-BEV-Tiny has 37.2\% NDS with only 15.8 MB model size, which outperforms the 8$\times$ larger BevFormer-Tiny model by 1.8\%.
\end{itemize}

\section{Related Works}
\label{sec:Relate}
\subsection{Camera-only 3D object detection}
In camera-only 3D object detection tasks, many excellent methods have emerged based on BEV (bird-eye-view).
Previous works, LSS~\cite{philion2020lift} and BEVDet~\cite{huang2021bevdet}, project image features to the BEV space in a bottom-up manner.
Based on DETR~\cite{carion2020end} and Deformable DETR~\cite{zhu2020deformable}, DETR3D~\cite{wang2022detr3d} extends the 2D object detection to the 3D space through the architecture of Backbone + FPN + Decoder. 
In addition, PETR~\cite{liu2022petr} introduces 3D position coding based on DETR3D~\cite{wang2022detr3d}. 
In BEVFormer~\cite{li2022bevformer}, the authors use dense BEV queries to exchange information in the BEV space with the multi-view image space. More stable BEV features are obtained by extracting temporal and spatial information through the transformer structure with temporal self-attention and spacial cross-attention. 
Based on the temporal interaction in BEVFormer, further improvements have been made in recent works, PETRv2~\cite{liu2022petrv2} and BEVDet4D~\cite{huang2022bevdet4d}. 
Besides the aforementioned works, BEVDepth~\cite{li2022bevdepth} and BEVstereo~\cite{li2022bevstereo} are two state-of-the-art approaches for monocular depth estimation and stereo vision in the bird-eye-view (BEV) domain, respectively, which leverage the unique characteristics of BEV representations to achieve high accuracy and efficiency.

\subsection{Quantization}
To reduce the model size, quantization~\cite{zhou2017incremental, wang2018haq, huang2021codenet, liu2023noisyquant} uses low bitwidth to represent weights and activations in neural networks.
With the use of low-precision matrix multiplication or convolution, quantization can also make the inference process faster and more efficient.
Given a pretrained model, directly performing quantization without any fine-tuning is referred to as post-training quantization (PTQ)~\cite{cai2020zeroq, nagel2019data, wang2020towards, kim2023squeezellm}. 
Despite the merits, PTQ with low bitwidth still results in significant accuracy degradation. As such, Quantization-aware training (QAT) is proposed to train the model to better adapt to quantization. QAT methods~\cite{dong2019hawqv2, choi2018pact, yao2021hawq} are more costly compared to PTQ, but can potentially obtain higher accuracy.
Furthermore, in the cases with ultra-low quantization bitwidth (for example, 4-bit), even QAT cannot close the accuracy gap. A promising direction to solve this is to use mixed-precision quantization~\cite{zhou2017adaptive,wang2018haq, xiao2022csq}, where some sensitive layers are kept at higher precision to recover the accuracy. Though effective, the support for mixed-precision quantization on general-purpose machines (CPUs and GPUs) is currently immature and may lead to extra latency overhead.

Although standard quantization methods have shown great results on convolutional neural networks, recent works~\cite{liu2021post, zhao2022analysis} mention that it may suffer in other neural architectures such as transformers. The presence of both convolutional blocks and transformers in BEV networks makes them challenging to the traditional quantization methods.

\subsection{Distillation}
Model distillation~\cite{hinton2015distilling, mishra2017apprentice, li2017learning, polino2018model, ahn2019variational} generally uses a large model as the teacher to train a compact student model. Instead of using class labels during the training of the student model, the key idea is to leverage the soft probabilities produced by the teacher to guide the student's training.
Previous methods of distillation explore different knowledge sources (for example,~\cite{hinton2015distilling, li2017learning, park2019relational} use logits, aka the soft probabilities). The choices of teacher models are also studied, where~\cite{you2017learning, tarvainen2017mean} use multiple teacher models, while~\cite{crowley2018moonshine, zhang2019your} apply self-distillation without an extra teacher model. Other previous efforts apply distillation with different settings on different applications.
Regarding BEV networks, previous work \cite{chen2022bevdistill} tries to teach LiDAR information to camera-based networks through distillation, but the additional requirement for LiDAR data makes it infeasible in our pure camera-based setting. Besides, the existence of different types of losses in BEV networks makes standard distillation methods invalid. An arbitrary or sub-optimal combination of knowledge sources would also make the training unstable, perform badly, or even diverge. 

\section{Method}


\subsection{Overview of QD-BEV pipeline}
\label{subsec:overview of qd-bev}

This work aims to improve the efficiency of state-of-the-art BEV models. Starting with the widely used BEVFormer~\cite{li2022bevformer}, we apply a progressive quantization-aware training procedure 
in a stage-by-stage manner (details are introduced in ~\cref{sec:PQAT}). 
We further boost its stability and performance through a novel view-guided distillation process, which is highlighted in Figure~\ref{fig:intro}, where we use a floating-point teacher model to facilitate the learning of our quantized QD-BEV student model.
Specifically, the input multi-camera image is entered into the teacher and the student, respectively, and then the  Image Backbone and Image Neck parts of the network are used to extract the multi-camera image feature.  After the transformer part of the network, the BEV features are extracted, and the two parts of the teacher model and the student model are used to calculate image distill loss and BEV distill loss, respectively. The two distillation losses are then fused through the external parameters of the camera to achieve our unique view-guided distillation mechanism. 
We provide a detailed formulation of the view-guided distillation process in~\cref{subsec:vgd}.

\subsection{Quantization-aware training}
\label{sec:PQAT}

In symmetric linear quantization, the quantizer maps weights and activations into integers with a scale factor $S$. Uniformly quantizing to k bit can be expressed as:
\small
\begin{equation}
\begin{split}
\textit{S} =\frac{2|r_{\max}|}{2^{k}-1},\ 
\textit{q} =\textit{round}\left(\frac{r}{S}\right)
\end{split}
\end{equation}
\normalsize
where $r$ is the floating-point number being quantized, $|r_{\max}|$ is the largest absolute value in $r$, and $q$ is the quantized integer.
In this work, we conduct systematic experiments to analyze the performance of quantization on BEV networks. For PTQ, we apply the above quantization directly to the pre-trained models during the inference stage. For QAT, we utilize the straight-through estimator (STE)~\cite{bengio2013estimating} to define the forward and backward processes for the above quantization operations, and then we train the model to better adapt to quantization.
As previously mentioned in Sec.~\ref{sec:intro} and Sec.~\ref{sec:Relate}, considering that standard QAT may lead to divergence due to the characteristics of BEV models, we apply a stage-wise progressive QAT where we gradually reduce the weight precision in four stages (backbone, neck, encoder, and decoder) based on the design of BEVFormer~\cite{li2022bevformer}. The performance of this progressive QAT is illustrated in Figure~\ref{fig:progressive}. And we compare the effectiveness of the progressive QAT with standard QAT in Sec.~\ref{sec:qat results}.

\subsection{View-guided distillation}
\label{subsec:vgd}
Compared to traditional single-domain distillation methods, our approach exploits the complementary nature of both BEV and image domains, which provides different perspectives and capture different aspects of the scene. The BEV domain offers a top-down view, enabling accurate perception and recognition of the surrounding environment, such as the structure of the road, the location of vehicles, and lane markings. On the other hand, the image domain provides more realistic visual information, capturing rich scene details and color information.
In the following sections, we present details of VGD: the computation of the image feature distillation in~\cref{subsec:img_feature}, the BEV feature distillation in~\cref{subsec:bev_feature}, and the view-guided distillation combining the previous two distillation losses in~\cref{subsec:compute_vgd}.

\subsubsection{Image feature distillation}
\label{subsec:img_feature}

Given a pair of aligned teacher and student models,
We first compute element-wise distillation loss on image features.
We extract the image neck output as the image features to be distilled. 
To improve the smoothness of the distillation loss, we follow previous attempts~\cite{shu2021channel} to use a KL divergence-based distillation loss.
Specifically, we consider the flattened image features of the student and the teacher model as logits, which we convert into probability distribution via a softmax function with temperature $\phi_{\tau}$, as defined in~\cref{eq:softmax_ImgFeature_t}.
\small
\begin{equation}
	{\phi_{\tau}(x_i)} = \frac{e^{x_i/\tau}}{\sum_j e^{x_{j}/\tau}}
	\label{eq:softmax_ImgFeature_t}
\end{equation}
\normalsize
Then we calculate the KL divergence of each camera's output separately to obtain the image feature distillation loss, as in~\cref{eq:Loss_img}.
\small
\begin{equation}
    \begin{split}
    {\mathcal{L}_{img}} = \frac{\tau ^2}{{ B}\cdot { W}\cdot { H}\cdot { C}} \times \mathcal{D}_{KL} \left( \phi_{\tau}(F^{T}_{img}),\phi_{\tau}(F^{S}_{img}) \right)
	\label{eq:Loss_img}
    \end{split}
\end{equation}
\normalsize
where B stands for the batch size, W, H, C mean the width, height, and number of channels of the image features, respectively. $F^{T}_{img}$ and $F^{S}_{img}$ denote the image features of the teacher model and the student model.

\subsubsection{BEV feature distillation}
\label{subsec:bev_feature}


We first convert the BEV features of the student and teacher into probability distributions, following the same procedure as for image features. Then we compute the KL divergence for each point on the BEV features, as shown in~\cref{eq:Loss_BEV}.
\small
\begin{equation}
    \begin{split}
    {\mathcal{L}_{bev}} = \frac{\tau ^2}{{ B}\cdot { C}} \times {\rm \mathcal{D}_{KL}} \left( \phi_{\tau}(F^{T}_{bev}),\phi_{\tau}(F^{S}_{bev}) \right)
	\label{eq:Loss_BEV}
    \end{split}
\end{equation}
\normalsize
where B stands for the batch size, C means the number of channels of BEV features. 
$F^{T}_{bev}$ and $F^{S}_{bev}$ denote the BEV features of the teacher model and student model, respectively. We will get a loss with the shape of $[ H_{bev} \times W_{bev}, 1]$.

\subsubsection{View-guided distillation objective}
\label{subsec:compute_vgd}



In the first two sections, we obtained the loss of each camera on the image feature and the corresponding loss of each point on the BEV feature. 
On the nuScenes dataset, the camera external parameters are known, so we can obtain the distribution range of each camera corresponding to the BEV feature. Then we generate the BEV mask of views $M_{bev}$ that can be applied to the image feature, which is the same as defined in BEVFormer~\cite{li2022bevformer}.  
$M_{bev}$ is a tensor with four dimensions: number of cameras, batch size, BEV size $(H_{bev} \times W_{bev})$, and 3D Height, with binary values in each element. 
By calculating the average along the last dimension (3D Height), we can flatten the BEV mask $M_{bev}$ on the 2d plane with the BEV size $(H_{bev} \times W_{bev})$. 
Then the ${\mathcal{L}_{img}}$ calculated for each camera can be extended to the corresponding loss for each point on the BEV feature, which we refer to as 
${\widehat{\mathcal{L}}_{img}}$:
\small
\begin{equation}
\begin{split}
{\widehat{\mathcal{L}}_{img}} = {\mathcal{L}_{img}} \odot M_{bev}
\label{eq:masked img loss}
\end{split}
\end{equation}
\normalsize
where $\odot$ denotes the hadamard product.


Finally, we use ${\widehat{\mathcal{L}}_{img}}$ to get the view-guided distillation objective in~\cref{eq:View Guided Loss}:
\small
\begin{equation}
\begin{split}
\mathcal{L}_{vgd} ={\sum_{i=1}^{N\cdot H\cdot W}} {\widehat{\mathcal{L}}_{img}} \odot {\mathcal{L}_{bev}}
\label{eq:View Guided Loss}
\end{split}
\end{equation}
\normalsize
The overall process of view-guided distillation is shown in Algorithm~\ref{algorithm1}.

\renewcommand{\algorithmicrequire}{\textbf{Input:}}
\renewcommand{\algorithmicensure}{\textbf{Output:}}

\renewcommand{\algorithmicrequire}{\textbf{Input:}}
\renewcommand{\algorithmicensure}{\textbf{Output:}}
\begin{algorithm}[t]
        \caption{\text{Progressive Quantization-Aware VGD}}
        \begin{algorithmic}[0]
            \Require Training data; 
            \\ Pre-trained FP weights for teacher model $T_{fp}$; 
            \\ Student model $S$; 
            \\ Numbers of epochs for each phase $N_1$, ... , $N_4$; $P_2$;
            \\ Quantization bit for weight and activation $B_w; B_a$
            \Ensure Trained low-bit student model $S_q$ 
        \State \textbf{Phase 1: Progressive Quantization-Aware Training} 
        \State Init $S_{0}$ with $T_{fp}$; Divide model into 4 $Parts$: 
                $\lbrace Backbone, Neck, Encoder, Decoder \rbrace$;
        \For{$i, module$ in $Parts$}{}
            \State $module$ = quantize ($module$; $B_w$, $B_a$)
            \State Init $S_{i}$ with $S_{i-1}$
            \For{Epoch = $N_{i-1}$ ,..., $N_i$}{ }
                \State Update $S_{q_i}$ by minimizing QAT loss 
            \EndFor
        \EndFor
        \State \textbf{Phase 2: View-Guided Distillation} 
        \State Init ${S}$ with $S_{4}$  
        \For{ Epoch = 1, ... , $P_2$} 
            \State Update $S$ by minimizing QAT loss and $\mathcal{L}_{vgd}$ in~\cref{eq:View Guided Loss}
        \EndFor
        \end{algorithmic}
\label{algorithm1}
\end{algorithm}

\section{Experiments}
\label{sec:Exp}

In this section, we first elaborate on the experimental settings, then we evaluate both PTQ and QAT methods on the BEV networks. 
Based on the analysis of these results, we propose QD-BEV to overcome shortcomings in standard PTQ and QAT, and we dedicatedly compare our results with previous works under different settings and constraints.
\subsection{Experimental settings}
\subsubsection{Dataset}


We evaluated our proposed method on a challenging 3D detection task using the nuScenes dataset \cite{caesar2020nuscenes}, which is a large-scale public dataset for autopilot developed by the team at Motional (formerly nuTonomy). This dataset contains 1000 manually selected 20-second driving scenarios collected in Boston and Singapore, with 750 scenarios for training, 100 for validation, and 150 for testing. The images in the dataset were captured from 6 cameras with known internal and external parameters.

\begin{table*}[!t]
\tiny
\renewcommand\arraystretch{1}
    \centering
    \caption{Results with  PTQ separately applied on each part in BEVFormer.}
    \label{tab: part Result}
\resizebox{1\linewidth}{!}{
\begin{tabular}{cccc|cc|cc|cc} 
\toprule
 \multicolumn{4}{c|}{Quantized Module}  & \multicolumn{2}{c|}{BEVFormer-Tiny} & \multicolumn{2}{c|}{BEVFormer-Small} & \multicolumn{2}{c}{BEVFormer-Base} \\
\midrule
Backbone& Neck& Encoder& Decoder & mAP↑ & NDS↑ & mAP↑ & NDS↑ & mAP↑ & NDS↑ \\
\midrule 
\fontsize{4pt}{24pt}\XSolid & \fontsize{4pt}{24pt}\XSolid & \fontsize{4pt}{24pt}\XSolid & \fontsize{4pt}{24pt}\XSolid & 0.252 & 0.354 & 0.370 & 0.479 & 0.416 & 0.517\\

\midrule
\fontsize{5pt}{24pt}\CheckmarkBold & \fontsize{4pt}{24pt}\XSolid & \fontsize{4pt}{24pt}\XSolid & \fontsize{4pt}{24pt}\XSolid & 0.212 & 0.310 & 0.288 & 0.417 & 0.309 & 0.440\\

\fontsize{4pt}{24pt}\XSolid & \fontsize{5pt}{24pt}\CheckmarkBold & \fontsize{4pt}{24pt}\XSolid & \fontsize{4pt}{24pt}\XSolid & \bf{0.252} & \bf{0.353} & \bf{0.370} & \bf{0.478} & \bf{0.416} & 0.516  \\

\fontsize{4pt}{24pt}\XSolid & \fontsize{4pt}{24pt}\XSolid & \fontsize{5pt}{24pt}\CheckmarkBold & \fontsize{4pt}{24pt}\XSolid & 0.181 & 0.295 & 0.298 & 0.421 & 0.160 & 0.303\\

\fontsize{4pt}{24pt}\XSolid & \fontsize{4pt}{24pt}\XSolid & \fontsize{4pt}{24pt}\XSolid & \fontsize{5pt}{24pt}\CheckmarkBold & 0.251 & \bf{0.353} & 0.369 & 0.477 & \bf{0.416} & \bf{0.517}\\
\midrule
\fontsize{5pt}{24pt}\CheckmarkBold & \fontsize{5pt}{24pt}\CheckmarkBold & \fontsize{5pt}{24pt}\CheckmarkBold & \fontsize{5pt}{24pt}\CheckmarkBold & 0.145 & 0.246 & 0.228 & 0.369 & 0.075 & 0.224\\
\bottomrule
\end{tabular}
}
\end{table*}

\subsubsection{Evaluation metrics}

%
The main measurement indicators on the nuScenes 3D test dataset are the mean Average Precision (mAP), and the unique evaluation index nuScenes detection score (NDS). NDS is a comprehensive evaluation index containing many aspects of information. 
The other indicators are mean Average Translation Error (mATE), mean Average Scale Error (mASE), mean Average Orientation Error (mAOE), mean Average Velocity Error (mAVE) and mean Average Attribute Error (mAAE).
To evaluate the efficiency of the BEV networks, we use model size and BOPS as the metrics. 
Model size is the memory required to store a specific network, which is determined by the total amount of parameters in the model as well as the quantization bitwidth to store those parameters.
BOPS measures the total Bit Operations of one network inference~\cite{van2020bayesian}. It is a common metric for evaluating the computation of quantized neural networks. For a model with $L$ layers, defining $b_{w_i}$ and $b_{a_i}$ to be the bitwidth used for weights and activations of the $i$-th layer, then we have:
\small
\begin{equation}
\begin{split}
\textit{BOPS} = \sum\nolimits_{i=1}^L b_{w_i}b_{a_i} \textit{MAC}_i
\end{split}
\end{equation} 
\normalsize
where MAC$_i$ is the total Multiply-Accumulate operations for computing the $i$-th layer.
To better demonstrate the advantages of our QD-BEV model over the floating-point model, we introduce sAP~\cite{li2020towards} (streaming average precision) as one of the evaluation metrics for our model performance. sAP is a dynamic metric that will be updated as new data arrive, making it ideal for evaluating models in real-time scenarios.

\subsubsection{Baselines \& Implementation Details}


We mainly compare with floating-point baseline models proposed by BEVFormer~\cite{li2022bevformer} with different input image resolutions (specifically, BEVFormer-Tiny, BEVFormer-Small, and BEVFormer-Base).
For quantization baselines, we apply the previous PTQ method DFQ~\cite{nagel2019data} as well as QAT method PACT~\cite{choi2018pact} and HAWQv3~\cite{yao2021hawq} to quantize BEVFormer models and compare with QD-BEV.



For floating-point models, we use the open-sourced repository of BEVFormer with two different backbones: ResNet50 and ResNet101-DCN. 
We use symmetrical linear quantization in a channel-wise manner for weights, and layer-wise quantization for activations, which are the standard settings for previous quantization methods. 
Since there is little related work in our settings for reference, we make our training adopt the same scheme as the original training strategy of BEVFormer, which is to train 24 epochs in each step, and to use optimizer of AdamW, a starting lr as 2e-4, a linear warmup of 500 iters, and cosine annealing.




\begin{table}[h]
\caption{PTQ results with different quantization bitwidth.}
\label{table:ptq results}
\centering
\resizebox{1\linewidth}{!}{
\begin{tabular}{c | c c c c c}
\toprule
W-bit/A-bit            & Model   & NDS↑  & NDS Drop  &  mAP↑    \\ 
\hline
\multirow{3}{*}{32/32}  & Tiny   & 0.354 &     -     & 0.252 \\ 
                        & Small  & 0.479 &     -     &  0.370  \\
                        & Base   & 0.517 &     -     &  0.416  \\
\midrule
\multirow{3}{*}{8/8}    & Tiny   & 0.351 &   0.8\%    &  0.248  \\ 
                        & Small  & 0.477 &   0.4\%    &  0.366  \\
                        & Base   & 0.487 &   5.8\%    &  0.384  \\
\midrule
\multirow{3}{*}{6/6}    & Tiny   & 0.312 &   11.9\%    &  0.203  \\ 
                        & Small  & 0.430 &   10.2\%    &  0.306  \\
                        & Base   & 0.402 &   22.2\%    &  0.262  \\
\midrule
\multirow{3}{*}{4/6}    & Tiny   & 0.246 &   30.5\%    &  0.146  \\ 
                        & Small  & 0.369 &   23.0\%    &  0.228  \\
                        & Base   & 0.226 &   56.3\%    &  0.076  \\
\midrule
\multirow{3}{*}{4/4}    & Tiny   & 0.034 &   90.4\%    &  0.001  \\ 
                        & Small  & 0.034 &   92.9\%    &  0.001  \\
                        & Base   & 0.023 &   95.6\%    &  0.000  \\
\bottomrule
\end{tabular}}
\end{table}

\begin{table}[h]
\caption{Comparsion between progressive and standard QAT with W4A6. Please refer to Table~\ref{table:ptq results} for baseline FP accuracy (W32A32).}
\centering
\label{table:progressive constract}
\resizebox{1\linewidth}{!}{
\begin{tabular}{c|ccccc}
\toprule

Method & Model & NDS↑ & NDS Drop & mAP↑ \\
\midrule
\multirow{3}{*}{Standard QAT} & Tiny & 0.326 & 7.9\% & 0.216  \\
& Small & 0.421 & 12.1\% & 0.303 \\
& base  & 0.224 & 56.7\% & 0.071 \\
\midrule
\multirow{3}{*}{Progressive QAT} & Tiny & 0.348 & 1.7\% & 0.234\\
& Small & 0.467 & 2.5\% & 0.356 \\
& base  & 0.485 & 6.2\% & 0.376 \\

\bottomrule
\end{tabular}}
\end{table}

\subsection{Analysis on BEV Quantization}

\begin{table*}[!t]
\scriptsize
\renewcommand\arraystretch{1.5}
    \centering
    \caption{QD-BEV results compared to previous methods or baselines.}
    \label{tab: sota all}
\resizebox{1\linewidth}{!}
{
\begingroup
    \setlength{\tabcolsep}{2pt}
\begin{tabular}{c|c|cc|cc|ccccc} 
\toprule
Input Size & Model & Model Size(MB) & BOPS(Tera) & NDS↑  & mAP↑  & mATE↓  & mASE↓  & mAOE↓  & mAVE↓  & mAAE↓   \\ 
\midrule
\multirow{5}{*}{450$\times$800}  & BEVFormer-T\cite{li2022bevformer} & 126.8  & 62.33 & 0.354 & 0.253 & 0.899  & 0.294  & 0.655  & 0.657  & 0.216   \\
                                 & BEVFormer-T-DFQ\cite{nagel2019data}  & 31.7 & 3.90 & 0.340 & 0.236 & 0.949  & 0.296  & 0.651  & 0.671  & 0.217   \\
                                 & BEVFormer-T-HAWQv3\cite{yao2021hawq} & 15.9 & 1.46 & 0.348 & 0.234 & 0.949 & 0.304 & 0.568 & 0.661 & 0.209   \\
                                 & BEVFormer-T-PACT\cite{choi2018pact} & 15.9 & 1.46 & 0.347 & 0.234 & 0.919 & 0.289 & 0.604 & 0.671 & 0.216   \\
                                 \rowcolor{orange!40}\cellcolor{orange!0}& QD-BEV-T (Ours)  & 15.9 & 1.46 & 0.372 & 0.255 & 0.882  & 0.321  & 0.543  & 0.599  & 0.214   \\ 
\midrule
\multirow{5}{*}{720$\times$1280} & BEVFormer-S\cite{li2022bevformer} & 225.6 & 236.13 & 0.479 & 0.370 & 0.722 & 0.279 & 0.407 & 0.438 & 0.220 \\
                                 & BEVFormer-S-DFQ\cite{nagel2019data} & 56.4 & 14.76 & 0.467 & 0.356 & 0.751  & 0.283  & 0.405  & 0.451  & 0.222 \\
                                 & BEVFormer-S-HAWQv3\cite{yao2021hawq} & 28.2 & 5.53 & 0.467 & 0.356 & 0.751 & 0.287 & 0.419 & 0.449 & 0.208 \\
                                 & BEVFormer-S-PACT\cite{choi2018pact} & 28.2 & 5.53 & 0.461 & 0.351 & 0.750 & 0.275 & 0.422 & 0.507 & 0.196 \\
                                 \rowcolor{orange!40}\cellcolor{orange!0}& QD-BEV-S (Ours) & 28.2 & 5.53 & 0.479 & 0.374 & 0.716 & 0.290 & 0.389 & 0.474 & 0.210 \\ 
\midrule
\multirow{5}{*}{900$\times$1600} & BEVFormer-B\cite{li2022bevformer} & 262.9 & 667.39 & 0.517 & 0.416 & 0.672 & 0.273 & 0.370  & 0.393  & 0.197 \\
                                 & BEVFormer-B-DFQ\cite{nagel2019data} & 65.7 & 41.71 & 0.486 & 0.384 & 0.771  & 0.274  & 0.378 & 0.427 & 0.206 \\
                                 & BEVFormer-B-HAWQv3\cite{yao2021hawq} & 32.9 & 15.64 & 0.485 & 0.376 & 0.727 & 0.288 & 0.381 & 0.434 & 0.202 \\
                                 & BEVFormer-B-PACT\cite{choi2018pact} & 32.9 & 15.64 & 0.480 & 0.374 & 0.735 & 0.291 & 0.392 & 0.458 & 0.201 \\
                                 \rowcolor{orange!40}\cellcolor{orange!0} & QD-BEV-B (Ours) & 32.9 & 15.64 &  0.509 & 0.406 & 0.691 & 0.285 & 0.360 & 0.410 & 0.190 \\
\bottomrule
\end{tabular}
\endgroup 
}
\end{table*}


\begin{table}[htbp]
\scriptsize
\renewcommand\arraystretch{1.5}
    \centering
    \caption{\small{QD-BEV results on BEVDepth\cite{li2022bevdepth} and PETR\cite{liu2022petr}.}}
    \label{tab: BEVDepth table}
\resizebox{1\linewidth}{!}{
\begin{tabular}{cc|c|cc} 
\toprule
W & A & Model & mAP↑ & NDS↑ \\ 
\midrule
32 & 32 & BEVDepth-T\cite{li2022bevdepth} & 0.330 & 0.435 \\
\midrule
8 & 8 & BEVDepth-T-DFQ\cite{nagel2019data} & 0.281 & 0.377 \\
\midrule
4 & 6 & BEVDepth-T-HAWQ\cite{yao2021hawq} & 0.136 & 0.206  \\
\midrule
4 & 6 & BEVDepth-T-PACT\cite{choi2018pact}  & 0.270 & 0.362  \\
\midrule
\rowcolor{orange!40} 4 & 6 & QD-BEVDepth-T (Ours) & 0.301 & 0.394 \\
\midrule
32 & 32 & PETR-r50dcn\cite{liu2022petr} & 0.317 & 0.366 \\
\midrule
8 & 8 & PETR-r50dcn-DFQ\cite{nagel2019data} & 0.290 & 0.343 \\
\midrule
4 & 6 & PETR-r50dcn-HAWQ\cite{yao2021hawq}  & 0.162 & 0.216 \\
\midrule
4 & 6 & PETR-r50dcn-PACT\cite{choi2018pact}   & 0.268 & 0.320 \\
\midrule
\rowcolor{orange!40} 4 & 6 & QD-PETR-r50dcn (Ours) & 0.288 & 0.334 \\
\bottomrule
\end{tabular}
}
\vspace{-10pt}
\end{table}

\subsubsection{PTQ results}


We first analyze the sensitivity of different modules to quantization in Table~\ref{tab: part Result}. It can be seen that the backbone and the encoder parts of the network are more sensitive, while quantization of the neck and the decoder parts only brings a slight disturbance to the accuracy. We want to note that, based on the sensitivity analysis, it is possible to apply mixed-precision quantization to better preserve the sensitive modules, but we leave this as future work since it is beyond the scope of this paper.

We then analyze the influence of different quantization bitwidth on the final performance.
In Table \ref{table:ptq results}, directly applying PTQ with less than 8-bit precision will lead to a significant accuracy drop, especially when quantized to W4A4 the results become pure noise with around 0 mAP. As can be observed from Table~\ref{table:ptq results}, performing QAT is necessary in order to preserve the accuracy while achieving ultra-low bit quantization. 

\subsubsection{QAT results}
\label{sec:qat results}

\begin{figure*}[htp]
\centering
      \includegraphics[width=1\linewidth, trim={0 50pt 0 0}, clip]{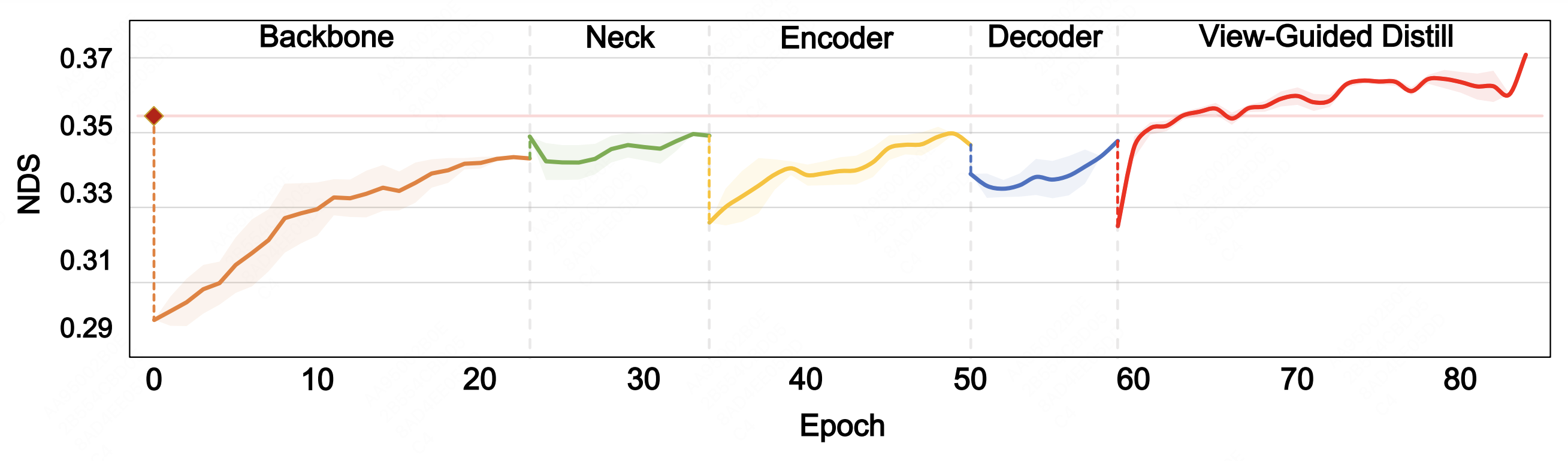}
   \caption{The training process of QD-BEV. Progressive QAT with 4 stages is performed in the first 60 epochs, then the view-guided distillation is conducted together with QAT to steadily enhance the performance. The pink line in the figure represents the baseline NDS. Note that swapping VGD with pure QAT or other distillation methods (for example, CWD~\cite{shu2021channel}) will lead to undesirable results, as we show in the last 20 epochs in this figure, compared with Figure~\ref{fig:vgd curve} and Figure~\ref{fig:cwd-vs-vgd}.
   }
   \label{fig:progressive}
\end{figure*}




To address the severe accuracy degradation of PTQ, we apply QAT to better adapt the models towards 4-bit quantization.
In all our experiments, the standard QAT method which directly quantizes the whole network to the target bitwidth will lead to unstable QAT processes, causing gradient explosion or a rapid decrease in accuracy in large models (for example, the W4A6 BEVFormer-Base has only 0.07 mAP). 
Based on this observation, we hypothesize that the quantization perturbation introduced in standard QAT is too large to be recovered. As such, we apply progressive QAT to constrain the quantization perturbation along the training process.
A comparison of the performance between progressive QAT and standard QAT with the same number of training epochs is presented in Table~\ref{table:progressive constract}. We can see that progressive QAT consistently outperforms standard QAT by a large margin (up to 5\% mAP) in BEVFormer-Tiny and BEVFormer-Small, and achieves an even larger performance gain on BEVFormer-Base.
To better validate our analyses, we plot the training curve of progressive QAT as the first 60 epochs in Figure \ref{fig:progressive}, where W4A6 quantization is conducted on BEVFormer-Tiny. We separate the progressive QAT into 4 stages and iteratively quantize a new module in each stage. As can be seen, there is an NDS drop at beginning of each stage, corresponding to the quantization perturbation introduced by quantizing each new module. 

Despite the merits of progressive QAT, we want to note that it still suffers from unstable training and performance degradation. As illustrated in Figure~\ref{fig:vgd curve} which corresponds to the last 20 epochs in Figure~\ref{fig:progressive}, progressive QAT keeps going up and down after reaching a plateau, while training curves of VGD show a rising trend with much better stability.

\subsection{Main Results of QD-BEV}


In order to obtain better accuracy and stability, we apply view-guided distillation with the floating-point model as the teacher and the quantized model as the student. The effect of VGD on W4A6 quantization of BEVFormer-Tiny is shown in Figure~\ref{fig:progressive}. Note that we separate VGD from progressive QAT in the first 60 epochs for a clearer comparison and illustration, and VGD is actually a plug-and-play function that can always be jointly applied with QAT, as we do in the last 20 epochs.
Benefiting from knowledge in both the image domain and the BEV domain, QD-BEV networks are able to fully recover the quantization degradation, and even outperform the floating-point baselines. As shown in Table~\ref{tab: sota all}, the NDS and mAP of the model outperform previous floating-point baselines as well as quantized networks. Since there are no existing results for compact BEV networks, we implement standard quantization methods DFQ~\cite{nagel2019data}, HAWQv3~\cite{yao2021hawq} and PACT~\cite{choi2018pact} on BEVFormer as a comparison. 
We apply W8A8 quantization for DFQ (DFQ is a PTQ method, lower bitwidth will lead to intolerable accuracy degradation) and W4A6 for QAT methods and QD-BEV models.
As a comparison, QD-BEV can achieve 0.509 NDS with only 32.9 MB model size, which is similar to the size of BEVFormer-T-DFQ (0.340 NDS) and much smaller than BEVFormer-Tiny (126.8 MB, 0.354 NDS).

Preliminary tests were run on PETR~\cite{liu2022petr} and BEVDepth~\cite{li2022bevdepth} models using our method in figure~\ref{tab: BEVDepth table}. Performance did not match BEVFormer, but still surpassed conventional quantization methods, highlighting the method's potential despite varying results.

\begin{figure}
    \centering
    \includegraphics[width=\linewidth, trim={0 0 0 0}, clip]{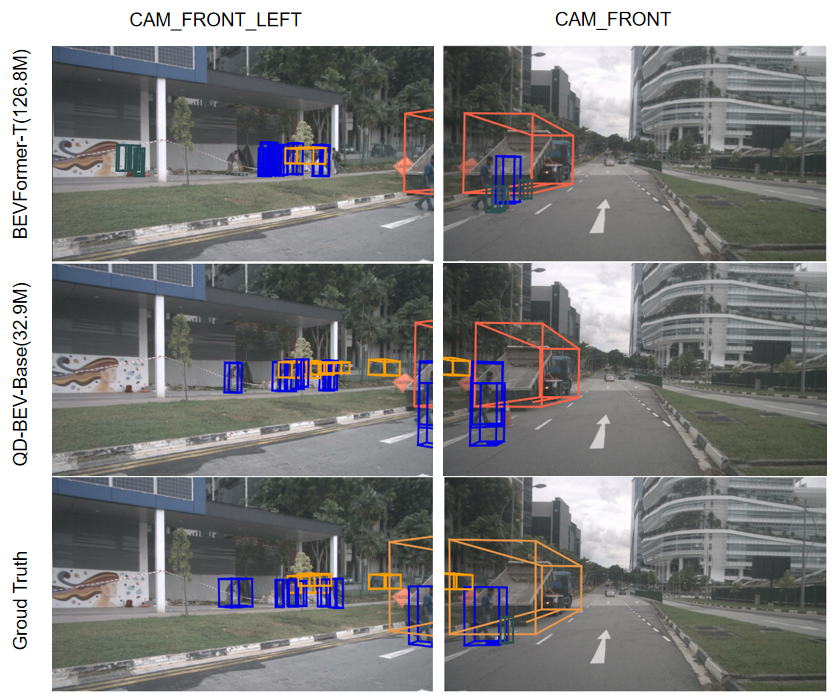}
    \caption{Visualization of QD-BEV results and the comparison with results obtained by BEVFormer.}
    \label{fig:visual1}
\end{figure}

In Figure~\ref{fig:visual1}, we show the visualization results of the QD-BEV-Base model on the nuScenes val dataset, and compare them with the results of BEVFormer-Tiny and the ground truth. 
As can be seen, more objects are detected by QD-BEV-Base, and the 3D boxes predictions are more accurate than BEVFormer-Tiny. More visualizations are provided in the supplementary material.


\subsection{Streaming perception result}
\label{sec: sAP}

In the context of autonomous driving, streaming perception~\cite{li2020towards} is critical for enabling models to make rapid and precise decisions in real time. High latency will degrade the streaming perception because it will cause a delay between the sensory data and the neural network output. To enhance streaming perception, quantization is a necessary technique that compresses the model size, reduces computational load, and accelerates the inference process. 
In Table~\ref{tab: sAP table}, we demonstrate the significant impact of quantization on the sAP metric in autonomous driving scenarios. As we can see, QD-BEV models show consistent improvements in sAP compared to the floating-point counterparts and the quantized baselines.

\begin{table}[!t]
\scriptsize
\renewcommand\arraystretch{1.5}
    \centering
    \caption{sAP of QD-BEV models compared to baselines.}
    \label{tab: sAP table}
\resizebox{1\linewidth}{!}
{
\begingroup
    \setlength{\tabcolsep}{2pt}
\begin{tabular}{c|c|cc|cccc} 
\toprule
Input Size & Model & W & A & mAP↑ & sAP↑\\ 
\midrule
\multirow{3}{*}{450$\times$800}  & BEVFormer-T\cite{li2022bevformer} & 32  & 32 & 0.253 & 0.228 \\
                                 & BEVFormer-T-DFQ\cite{nagel2019data}  & 8 & 8 & 0.236 & 0.230  \\
                                 \rowcolor{orange!40}\cellcolor{orange!0}& QD-BEV-T (Ours)  & 4 & 6  & 0.255 & \textbf{0.251} \\ 
\hline
\multirow{3}{*}{720$\times$1280} & BEVFormer-S\cite{li2022bevformer} & 32 & 32 & 0.370 & 0.249  \\
                                 & BEVFormer-S-DFQ\cite{nagel2019data} & 8 & 8 & 0.356 & 0.322  \\
                                 \rowcolor{orange!40}\cellcolor{orange!0}& QD-BEV-S (Ours) & 4 & 6 & 0.374 & \textbf{0.350} \\ 
\hline
\multirow{3}{*}{900$\times$1600} & BEVFormer-B\cite{li2022bevformer} & 32 & 32 & 0.416 & 0.136 \\
                                 & BEVFormer-B-DFQ\cite{nagel2019data} & 8 & 8 & 0.384 & 0.290 \\
                                 \rowcolor{orange!40}\cellcolor{orange!0} & QD-BEV-B (Ours) & 4 & 6 & 0.406 & \textbf{0.337} \\
\bottomrule
\end{tabular}
\endgroup 
}
\end{table}

\section{Ablation study}
\label{sec:ablation_study}

\begin{figure}
  \begin{subfigure}{0.495\linewidth}
    \includegraphics[width=4.3cm]{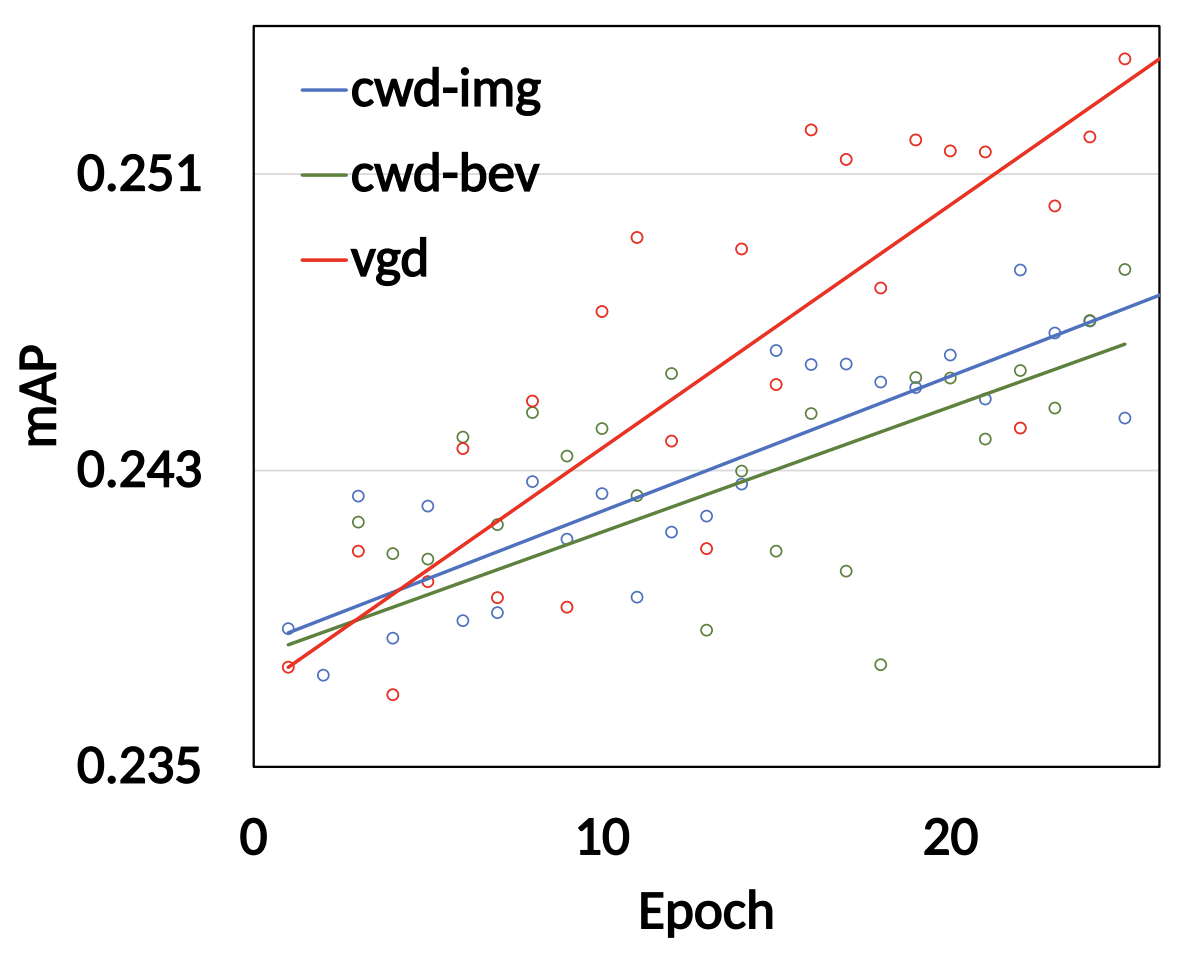}
    \caption{mAP curve.}
    \label{fig:cwd-vs-vgd_map_1}
  \end{subfigure}
  \begin{subfigure}{0.495\linewidth}
    \includegraphics[width=4.3cm]{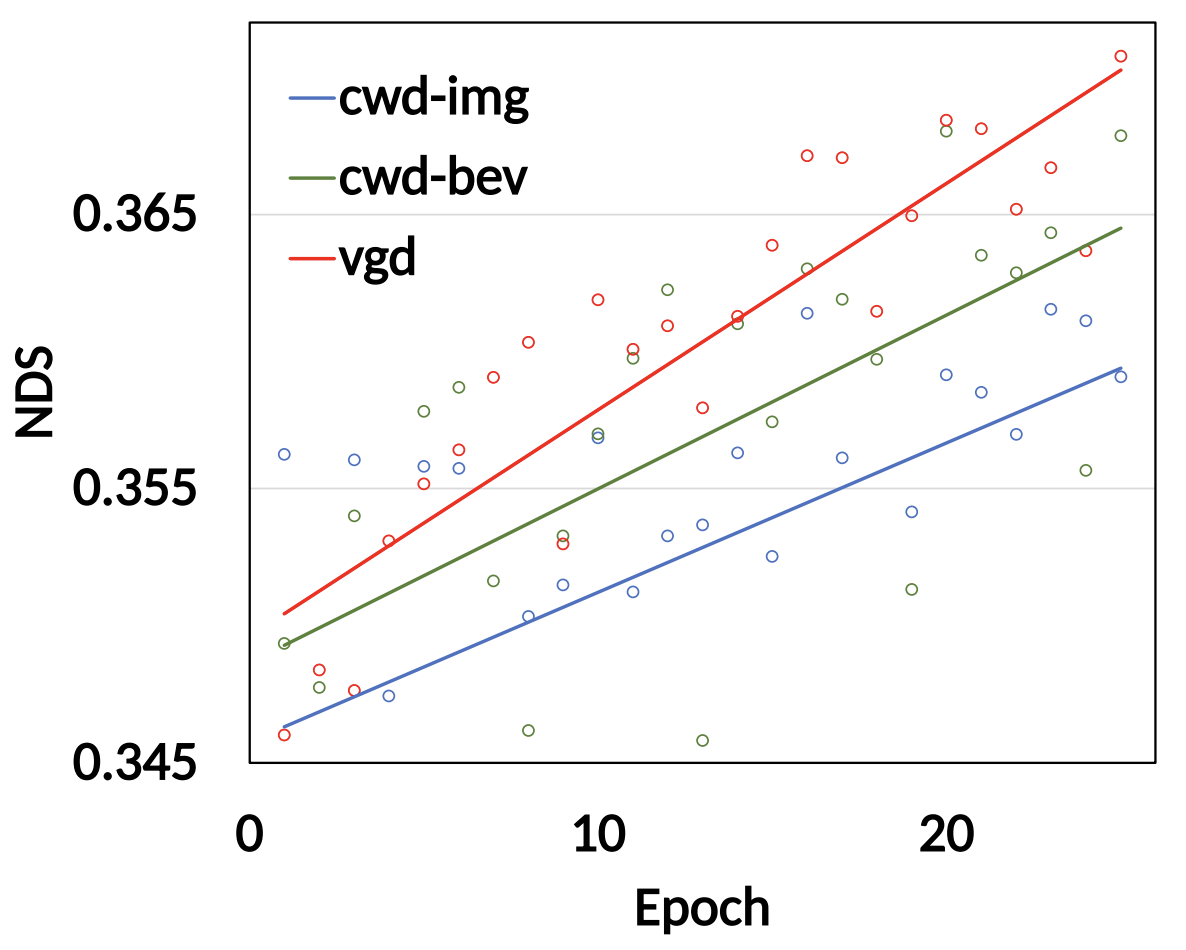}
    \caption{NDS curve.}
    \label{fig:cwd-vs-vgd_nds_1}
  \end{subfigure}
\caption{The training curve of VGD versus CWD~\cite{shu2021channel}. CWD uses only one type of features, referred to as cwd-img (image feature) and cwd-bev (bev feature), respectively.}

\label{fig:cwd-vs-vgd}
\end{figure}


In Figure \ref{fig:cwd-vs-vgd}, we conduct an ablation study where we compare view-guided distillation with the cases using only distillation on the image feature, and only on the BEV feature, respectively. The method is referred to as CWD~\cite{shu2021channel}.
For a fair comparison, we apply the same pre-training weights and hyperparameters such as the temperature and the learning rate. From the figure, it can be found that view-guided distillation has a clear advantage over the CWD methods. In both the mAP and NDS curves, VGD has a more obvious and stable upward trend, and achieves better final results.

\section{Conclusion}

In this work, we systematically study both PTQ and QAT on BEV networks and showcase the major problems they are facing. Based on our analyses, we propose a view-guided distillation (VGD) method that can stabilize the QAT process and enhance the final performance by leveraging information from both the image domain and the BEV domain. With VGD as a plug-and-play function that can be jointly applied when quantizing BEV models, QD-BEV can close the accuracy gap or even outperforms the floating-point baselines. On the nuScenes datasets, the 4-bit weight and 6-bit activation quantized QD-BEV-Tiny model achieves 37.2\% NDS with only 15.8 MB model size, outperforming BevFormer-Tiny by 1.8\% with an 8$\times$ model compression.
\section{Acknowledgement}
This work was supported by the National Key R\&D Program of China (2022ZD0116305). The authors would also like to express their gratitude to the CCF-Baidu Open Fund Project, Berkeley Deep Drive, and Intel Corporation for their support and assistance throughout this research.
Special thanks to Yandong Guo from AI\textsuperscript{2} Robotics (YANDONG.guo@live.com) for his valuable contributions to this work.

{\small
\bibliographystyle{ieee_fullname}
\bibliography{egbib}
}

\end{document}


\title{Supplementary Materials for QD-BEV}

\author{First Author\\
Institution1\\
Institution1 address\\
{\tt\small firstauthor@i1.org}
\and
Second Author\\
Institution2\\
First line of institution2 address\\
{\tt\small secondauthor@i2.org}
}
\maketitle

Our supplementary materials contain additional implementation details, ablation study, and visualization results.

\section{Additional Implementation Details}
\subsection{Supplementary description on datasets}
As we mentioned in the main body of the paper, the nuScenes \cite{caesar2020nuscenes} dataset has 750 scenarios as the training set, 100 scenarios as the validation set, and 150 scenarios as the test set. 
All our experiments are conducted on the nuScenes train set and tested on the nuScenes val set. 
Better results can be obtained using data enhancement and additional training, and some previous works \cite{wang2022detr3d,li2022bevformer,liu2022petrv2} use additional training data in order to get better results on the test set.
However, for the sake of fairness, we only train on the original training set and do not use techniques such as additional data and data enhancement.

\subsection{Extra details on training strategy}
Our experiments mainly use Tesla V100 32G GPU and Tesla A40 48G GPU to meet our video memory and computing power requirements. For the experiments of QD-BEV-Tiny, we use 8 pieces of Tesla A40 48G GPU with parallel computing, where the batch size is 6. For QD-BEV-Small and QD-BEV-Base experiments, we use 8 Tesla V100 GPU and 8 Tesla A40 GPU with parallel computing, where the batch size is 1. For the tiny, small, and base models, when the batch size is 1, the required single-card memory is 7G, 30G, and 47G, respectively.

For the training parameters, we generally follow the training configuration of the previous work\cite{li2022bevformer,wang2022detr3d}. In progressive quantization-aware training, we use the initial learning rate of 2e-4, learning rate multiplier of the backbone is 0.1 in each stage. But in view-guided distillation, we use an initial learning rate of 1e-5, and the learning rate multiplier of the backbone is 0.5.

For the temperature parameter $\tau$ of the view-guided distillation, our default configuration is $\tau$ = 1. 
To evaluate the sensitivity of the final performance to the hyperparameter $\tau$, we have done ablation experiments on our QD-BEV model with different $\tau$ in Section~\ref{subsec:temperature}.

\section{Additional Ablation Study}
\vspace{-5pt}
\subsection{Ablation study on the temperature parameter}
\label{subsec:temperature}
In the previous work\cite{shu2021channel}, it was found that the temperature parameter had an obvious effect on the results of distillation. 
Therefore, we carried out a control experiment with different hyperparameter $\tau$. 
We change the probability distribution of Softmax on the image feature and the BEV feature by selecting different $\tau$.




\begin{table}[htbp]
\tiny
\renewcommand\arraystretch{1.5}
    \centering
    \caption{Ablation study on the temperature parameter $\tau$ in VGD.}
    \label{tab: compare}
\centering
\resizebox{0.76\linewidth}{!}
{
\begin{tabular}{c|c|cc}
\toprule
Model                  & \fontsize{6pt}{6pt}{$\tau$}  & NDS   & mAP   \\
\midrule
\multirow{3}{*}{QD-BEV-Tiny}  & 1                    & 0.372 & 0.255 \\
                       & 2                    & 0.371 & 0.258 \\
                       & 4                    & {0.374} & {0.258} \\
\midrule
\multirow{2}{*}{QD-BEV-Small} & 1                   & 0.479 & 0.374 \\
                       & 4                    & 0.481 & 0.371 \\ 
\midrule
\multirow{2}{*}{QD-BEV-Base}  & 1                    & 0.506 & 0.403 \\
                       & 4                  & {0.509} & {0.406} \\
\bottomrule
\end{tabular}%
}
\end{table}

In Table~\ref{tab: compare}, we can see different hyperparameters do not have a decisive impact on the results. There is indeed some improvement in the performance of the three models when $\tau=4$, 
but the gap between a good result and a bad result is within 0.003 NDS, and the accuracies of all experiments are significantly and consistently higher than that of QAT. In addition, a possible reason for the 0.009 mAP difference in the QD-BEV-Base model is that the model has not been fully-trained. Due to time constraints, we have trained less than ten epochs in both configurations.

\section{Additional Visualization}

\begin{figure*}
\centering
      \includegraphics[width=0.86\linewidth]{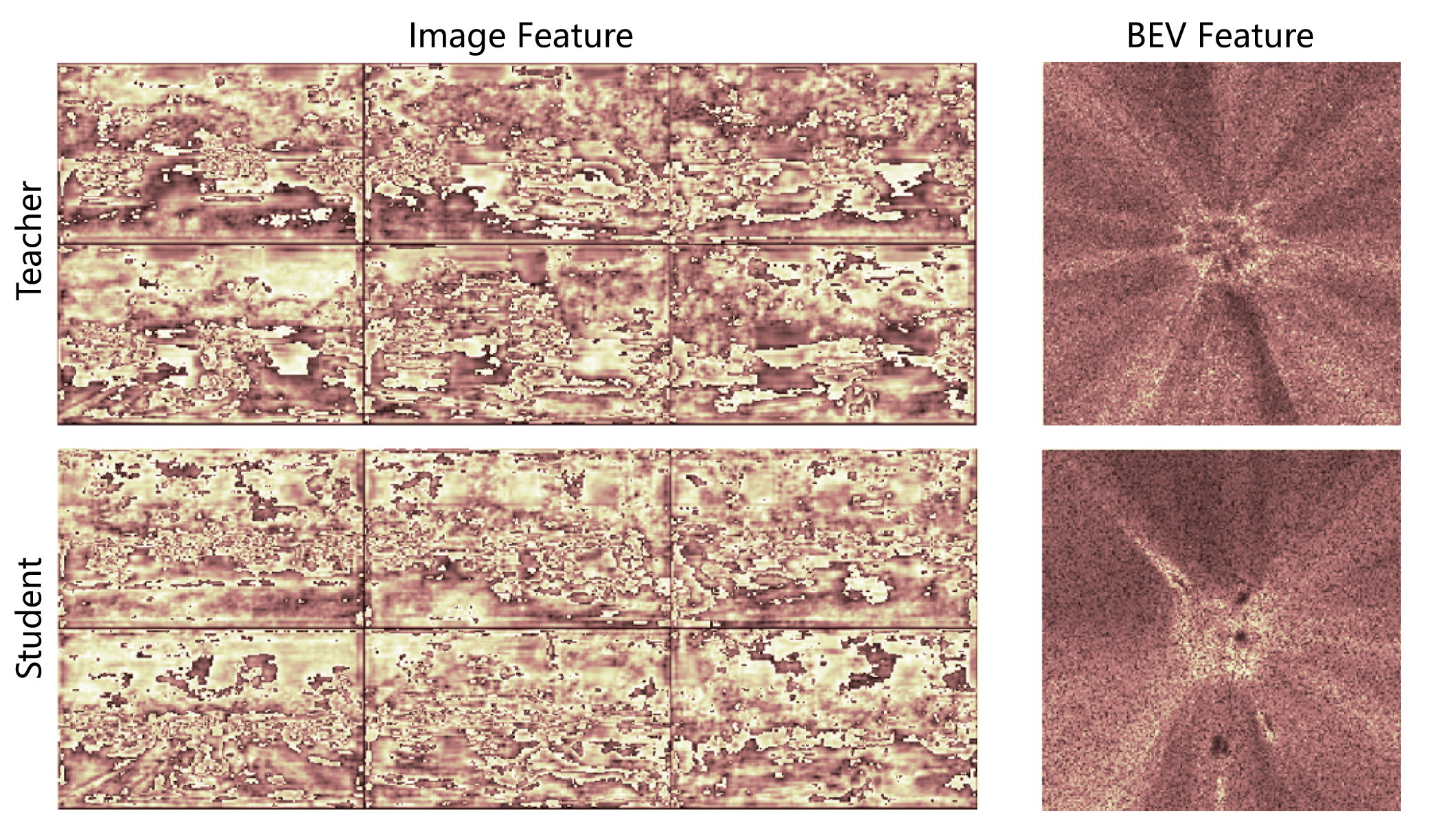}
   \caption{\textbf{Visualizaton of QD-BEV-Base feature map.} On the left is the Image features of the teacher and student model, while on the right we show the BEV features.}
   \label{fig:visusl_feature_map}
\end{figure*}
\subsection{Feature map visualization}

We visualize the feature map of the QD-BEV-Base model with view-guided distillation in Fig \ref{fig:visusl_feature_map}. The teacher is the BEVFormer-Base model, and the student is our QD-BEV-Base model with 4-bit weights and 6-bit activations. Our view-guided distillation method considers the Image feature and the BEV feature at the same time to obtain a better convergence direction of the model.

\subsection{Supplementary visualization on the oscillation of accuracy during QAT}
\begin{figure}
\centering
      \includegraphics[width=0.9\linewidth]{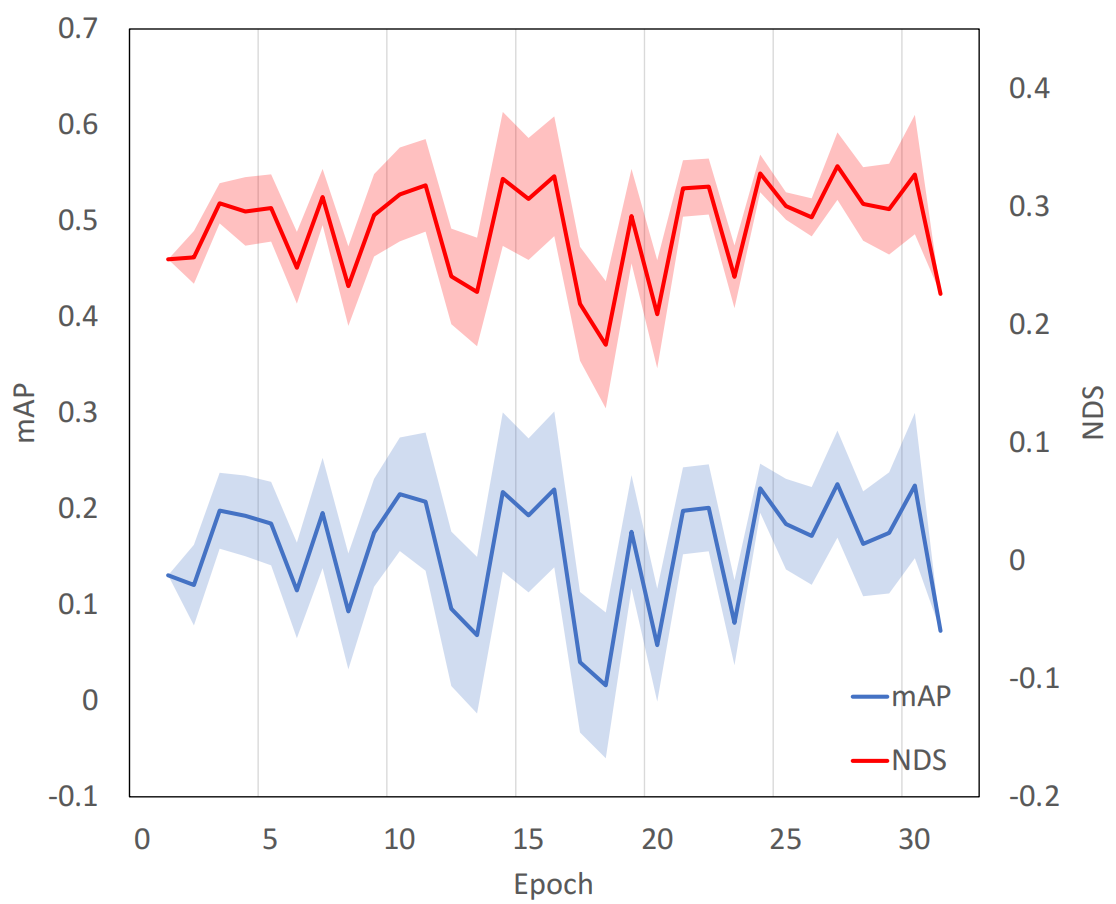}
   \caption{Visualization of the oscillation of accuracy during QAT.}
   \label{fig:noprogressive}
\end{figure}

During the quantization-aware training, we found that the accuracy curves of both mAP and NDS oscillate up and down throughout the process. We visualize this phenomenon in Fig \ref{fig:noprogressive}. We also discuss and demonstrate in the main texts that our progressive QAT with view-guided distillation method can help address this oscillation issue.

\subsection{Additional 3D object detection visualization}

\begin{figure*}[htbp]
  \centering
  \begin{subfigure}{1\linewidth}\centering
    \includegraphics[width=0.7\linewidth]{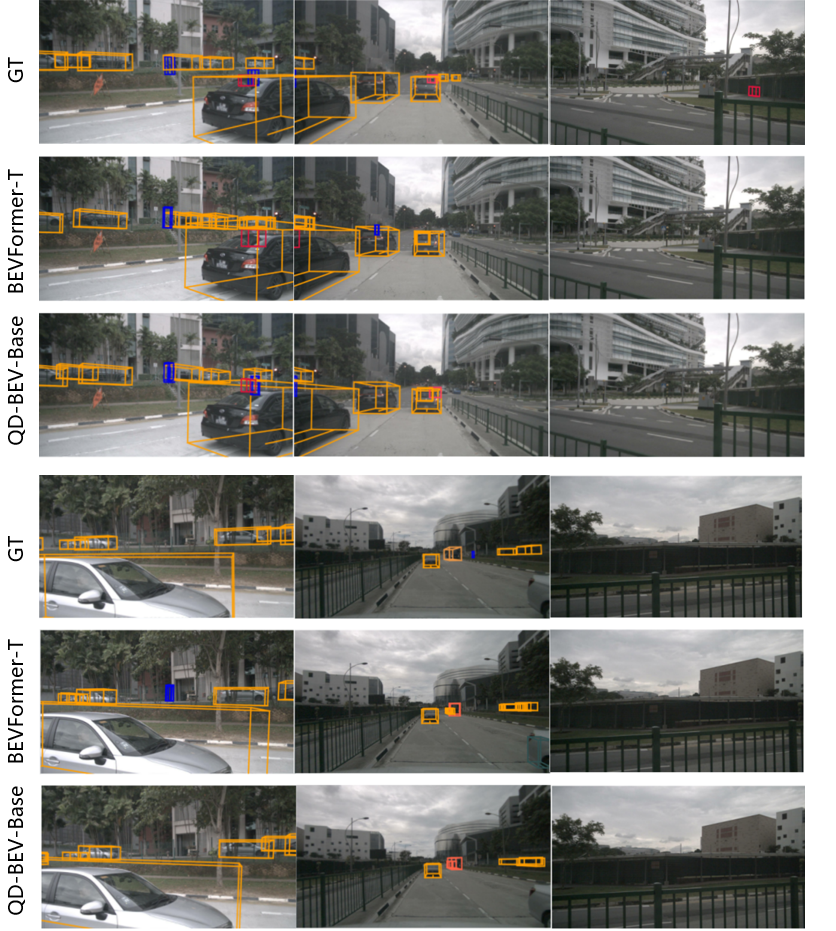}
    \caption{Visualization of 3D detection results of QD-BEV-Base, BEVFormer-T and Ground Truth}
    \label{fig:3d}
  \end{subfigure}
  \hfill
  \begin{subfigure}{1\linewidth}\centering
    \includegraphics[width=0.7\linewidth]{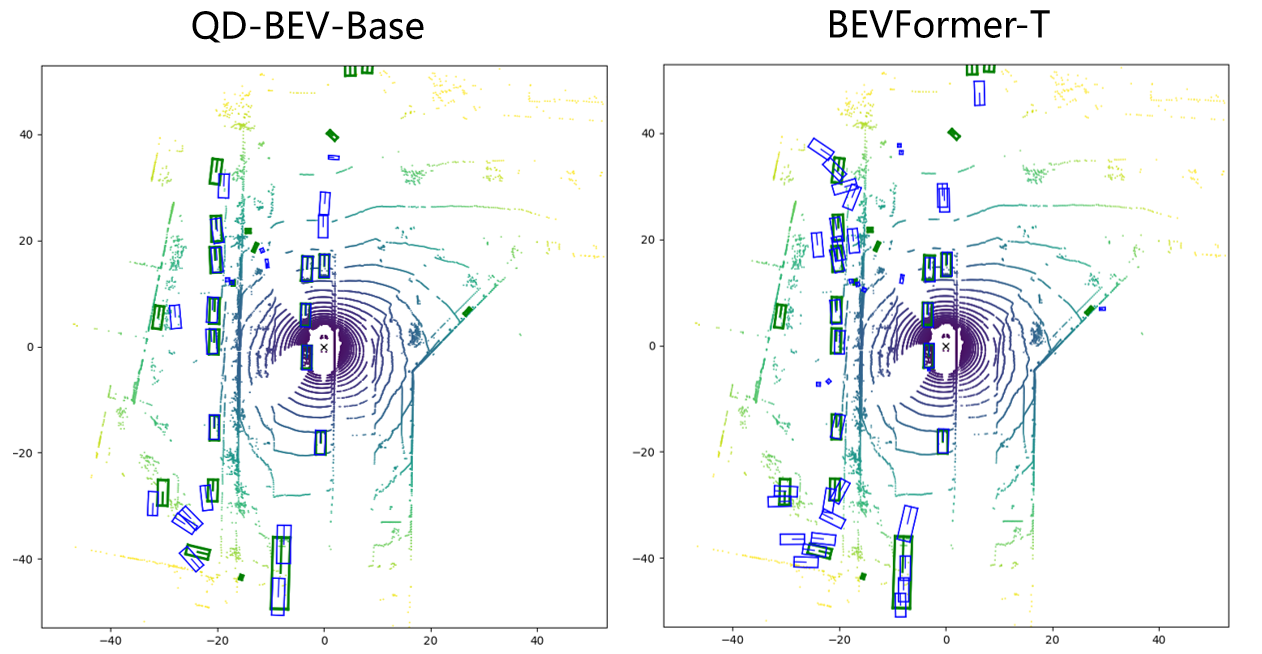}
    \caption{BEV visualization of QD-BEV-Base and BEVFormer-T}
    \label{fig:bev}
  \end{subfigure}
    \centering
  \caption{Visualization of QD-BEV-Base results and the comparison
with results obtained by BEVFormer-T and Ground Truth. The upper figures are from front cameras, and the lower figures are from back cameras.}
  \label{fig:Visual_sec7}
\end{figure*}

Here in Fig \ref{fig:Visual_sec7}, we visualize our model on more samples. 

In Fig \ref{fig:3d}, the upper part is the viewing angle of the three cameras in front of the car, and the lower half is the viewing angle of the three cameras behind the car. From top to bottom are the Ground Truth, BEVFormer-Tiny results, and our QD-BEV-Base results, respectively. We can find that QD-BEV-Base predicts the results more accurately than BEVFormer-T, and the error rate is also significantly lower.

In Fig \ref{fig:bev}, the BEV visualization of the QD-BEV-Base is on the left and the BEVFormer-Tiny result is shown on the right as a comparison. In these two pictures, the color blue represents the predicted result and the color green represents the Ground Truth. As can be seen in the figures, QD-BEV-Base (32.9MB) made a much clearer and more accurate prediction than BEVFormer-Tiny (126.8MB) with only 1/4 of the model size.

\section{Video Demo}

To showcase the efficacy of QD-BEV models, we made a 1080P HD video demo of QD-BEV-Base (32.9 MB) for about 20 seconds, including its comparison with BEVFormer-T (126.8 MB) and Ground Truth. 
The video is attached to the zip file.

{\small
\bibliographystyle{ieee_fullname}
\bibliography{latex/egbib_appendix}
}


\title{QD-BEV: Quantization-aware View-guided Distillation for Multi-view 3D Object Detection (Supplementary Materials)}

\author{First Author\\
Institution1\\
Institution1 address\\
{\tt\small firstauthor@i1.org}
\and
Second Author\\
Institution2\\
First line of institution2 address\\
{\tt\small secondauthor@i2.org}
}
\maketitle

\appendix

The supplementary materials contain additional implementation details, extra experimental results, ablation study, and visualization results.

\section{Additional Implementation Details}
\subsection{Supplementary description on datasets}
As we mentioned in the main body of the paper, the nuScenes \cite{caesar2020nuscenes} dataset has 750 scenarios as the training set, 100 scenarios as the validation set, and 150 scenarios as the test set. 
All our experiments are conducted on the nuScenes train set and tested on the nuScenes val set. 
Better results can be obtained using data enhancement and additional training, and some previous works \cite{wang2022detr3d,li2022bevformer,liu2022petrv2} use additional training data in order to get better results on the test set.
However, for the sake of fairness, we only train on the original training set and do not use techniques such as additional data and data enhancement.

\subsection{Extra details on training strategy}
Our experiments mainly use Tesla V100 32G GPU and Tesla A40 48G GPU to meet our video memory and computing power requirements. For the experiments of QD-BEV-Tiny, we use 8 pieces of Tesla A40 48G GPU with parallel computing, where the batch size is 6. For QD-BEV-Small and QD-BEV-Base experiments, we use 8 Tesla V100 GPU and 8 Tesla A40 GPU with parallel computing, where the batch size is 1. For the tiny, small, and base models, when the batch size is 1, the required single-card memory is 7G, 30G, and 47G, respectively.

For the training parameters, we generally follow the training configuration of the previous work~\cite{li2022bevformer,wang2022detr3d}. In progressive quantization-aware training, we use the initial learning rate of 2e-4, learning rate multiplier of the backbone is 0.1 in each stage. In view-guided distillation, we use an initial learning rate of 1e-5, and the learning rate multiplier of the backbone is 0.5.

For the temperature parameter $\tau$ of the view-guided distillation, our default configuration is $\tau$ = 1. 
To evaluate the sensitivity of the final performance to the hyperparameter $\tau$, we have conducted ablation study on our QD-BEV models with different $\tau$ in Section~\ref{subsec:temperature}.

\section{Additional Experiments on BEVDepth}
Our major experiments are conducted on top of BEVFormer. In order to further validate the generalization ability of our proposed methods, we also include preliminary experimental results on the BEVDepth model. From our experiments we found that, compared to BEVFormer, BEVDepth is more sensitive to quantization, and direct post-training quantization (PTQ) or standard quantization-aware training (QAT) methods result in intolerable accuracy loss. As such, we apply QD-BEV on BEVDepth and present the results in Table~\ref{tab: BEVDepth table}. It can be observed that QD-BEV can achieve significant accuracy improvement compared to other methods. The mAP and NDS scores of our 4-bit quantized models even outperform the 8-bit quantized counterparts from DFQ by a large margin. 

\begin{table}[htbp]
\scriptsize
\renewcommand\arraystretch{1.5}
    \centering
    \caption{QD-BEV results on BEVDepth compared to baselines.}
    \label{tab: BEVDepth table}
\resizebox{0.85\linewidth}{!}
{
\begingroup
    \setlength{\tabcolsep}{2pt}
\begin{tabular}{c|cc|cccc} 
\toprule
 Model & W & A & mAP↑ & NDS↑\\ 
\midrule
BEVDepth-T\cite{li2022bevdepth} & 32  & 32 & 0.330 & 0.435 \\
\midrule
BEVDepth-T-DFQ\cite{nagel2019data}  & 8 & 8 & 0.281 & 0.377  \\
\midrule
BEVDepth-T-HAWQ\cite{yao2021hawq}  & 4 & 6 & 0.136 & 0.206  \\
\midrule
BEVDepth-T-Progressive-QAT  & 4 & 6 & 0.275 & 0.369  \\
\midrule
 \rowcolor{orange!40} QD-BEVDepth-T (Ours)  & 4 & 6  & 0.301 & 0.394 \\ 
\bottomrule
\end{tabular}
\endgroup 
}
\vspace{-5pt}
\end{table}




\begin{table}[htbp]
\tiny
\renewcommand\arraystretch{1.5}
    \centering
    \caption{Ablation study on the temperature parameter $\tau$ in VGD.}
    \label{tab: compare}
\centering
\resizebox{0.76\linewidth}{!}
{
\begin{tabular}{c|c|cc}
\toprule
Model                  & \fontsize{6pt}{6pt}{$\tau$}  & NDS   & mAP   \\
\midrule
\multirow{3}{*}{QD-BEV-Tiny}  & 1                    & 0.372 & 0.255 \\
                       & 2                    & 0.371 & 0.258 \\
                       & 4                    & {0.374} & {0.258} \\
\midrule
\multirow{2}{*}{QD-BEV-Small} & 1                   & 0.479 & 0.374 \\
                       & 4                    & 0.481 & 0.371 \\ 
\midrule
\multirow{2}{*}{QD-BEV-Base}  & 1                    & 0.506 & 0.403 \\
                       & 4                  & {0.509} & {0.406} \\
\bottomrule
\end{tabular}%
}
\end{table}

\section{Additional Ablation Study}
\label{subsec:temperature}
In previous work~\cite{shu2021channel}, it is found that the temperature parameter has an obvious effect on the results of distillation. 
Therefore, we carried out a control experiment with different hyperparameter $\tau$. 
We change the probability distribution of Softmax on the image feature and the BEV feature by selecting different $\tau$.

In Table~\ref{tab: compare}, we can see different hyperparameters do not have a decisive impact on the results. There is indeed some improvement in the performance of the three models when $\tau=4$, 
but the gap between a good result and a bad result is within 0.003 NDS, and the accuracies of all experiments are significantly and consistently higher than that of QAT. 


\begin{figure*}[htbp]
\centering
      \includegraphics[width=0.8\linewidth]{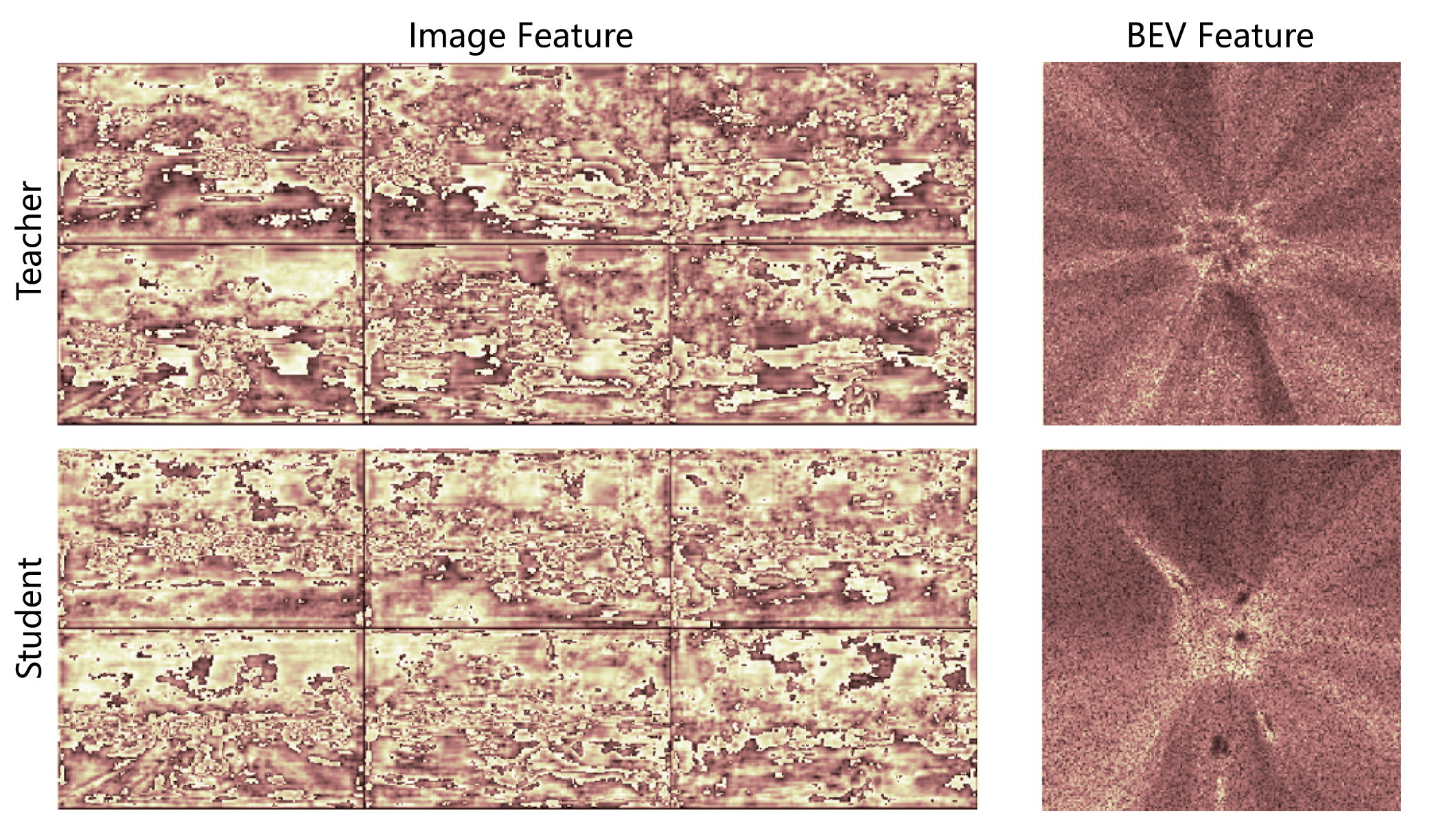}
   \caption{Visualizaton of QD-BEV-Base feature map. On the left is the Image features of the teacher and student model, while on the right we show the BEV features.}
   \label{fig:visusl_feature_map}
\end{figure*}

\begin{figure*}[htbp]
  \centering
  \begin{subfigure}{0.33\linewidth}\centering
    \includegraphics[width=1.1\linewidth]{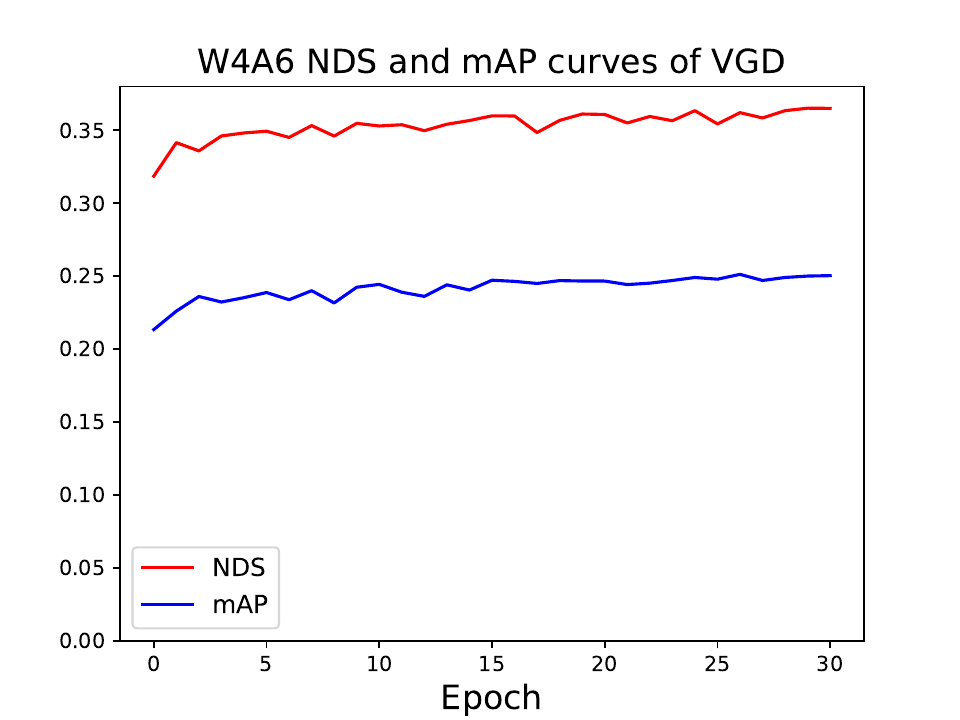}
    \caption{Oscillation of VGD}
    \label{fig:oscillation of VGD}
  \end{subfigure}
  \hfill
  \begin{subfigure}{0.33\linewidth}\centering
    \includegraphics[width=1.1\linewidth]{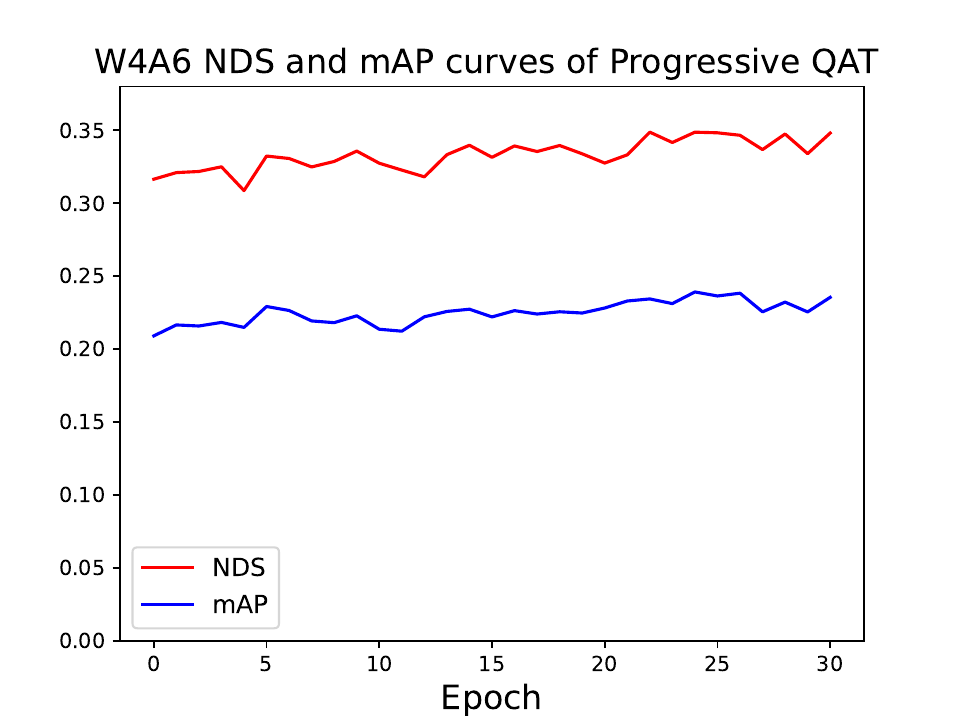}
    \caption{Oscillation of Progressive QAT}
    \label{fig:oscillation of Progressive QAT}
  \end{subfigure}
  \hfill
  \begin{subfigure}{0.33\linewidth}\centering
    \includegraphics[width=1.1\linewidth]{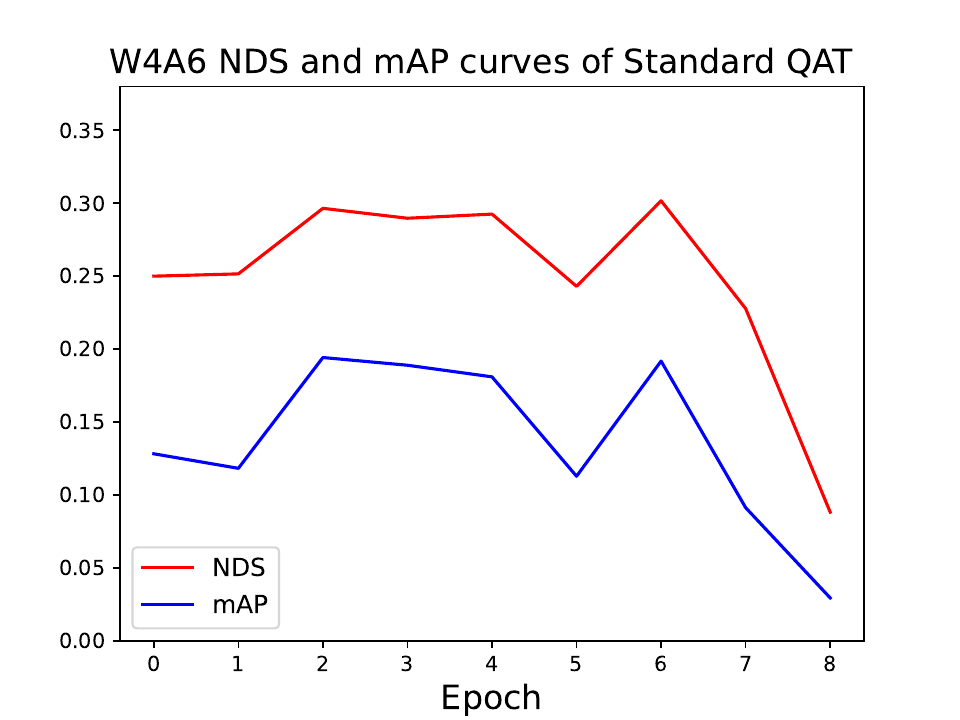}
    \caption{Oscillation of Standard QAT}
    \label{fig:oscillation of Standard QAT}
  \end{subfigure}
    \centering
  \caption{Visualization on the oscillation of accuracy during VGD, compared to Progressive QAT and Standard QAT.}
  \label{fig:oscillation}
\end{figure*}


\begin{figure*}[htbp]
  \centering
  \begin{subfigure}{0.33\linewidth}\centering
    \includegraphics[width=1.2\linewidth]{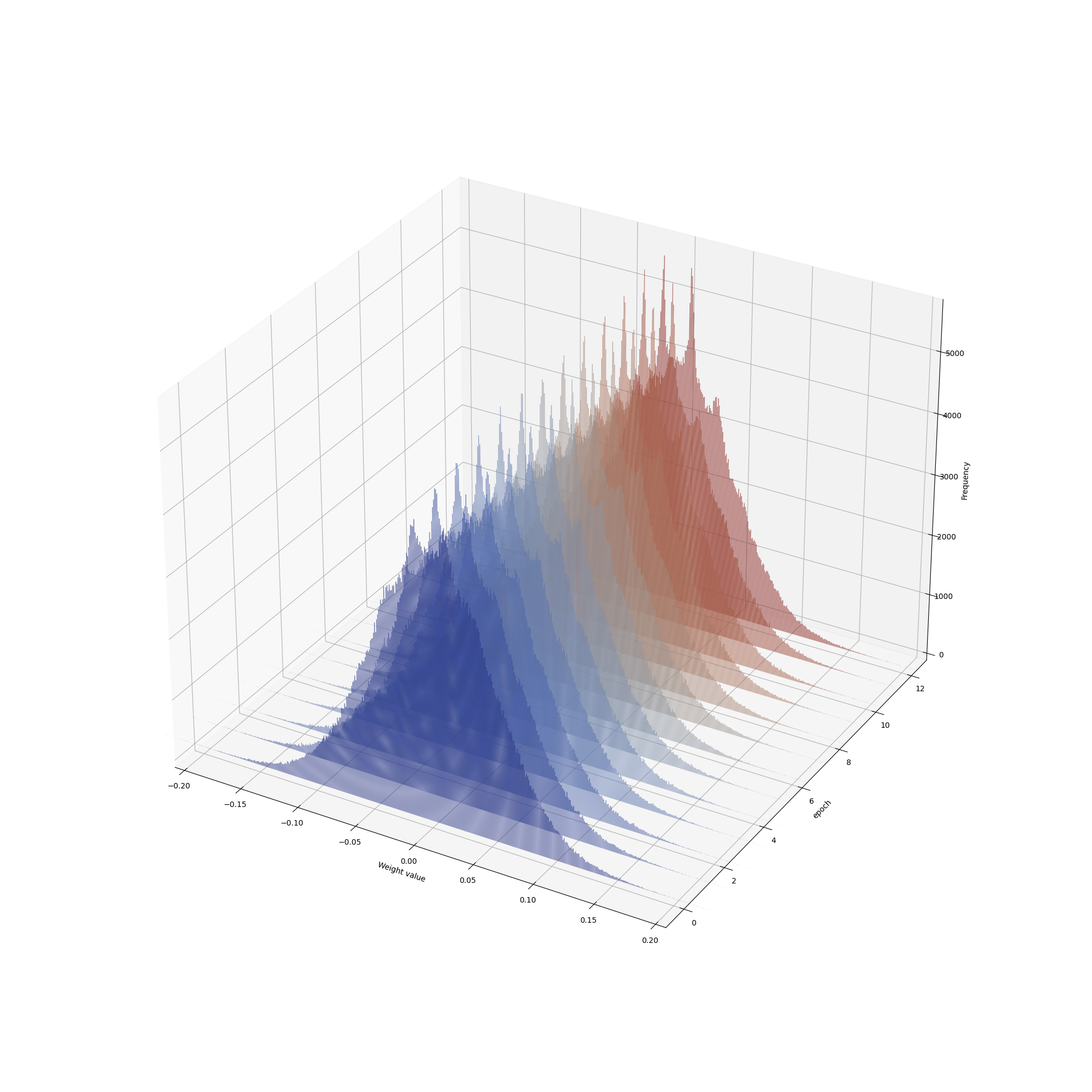}
    \caption{weight distribution of VGD}
    \label{fig:neck_qat_a4_ablation}
  \end{subfigure}
  \hfill
  \begin{subfigure}{0.33\linewidth}\centering
    \includegraphics[width=1.2\linewidth]{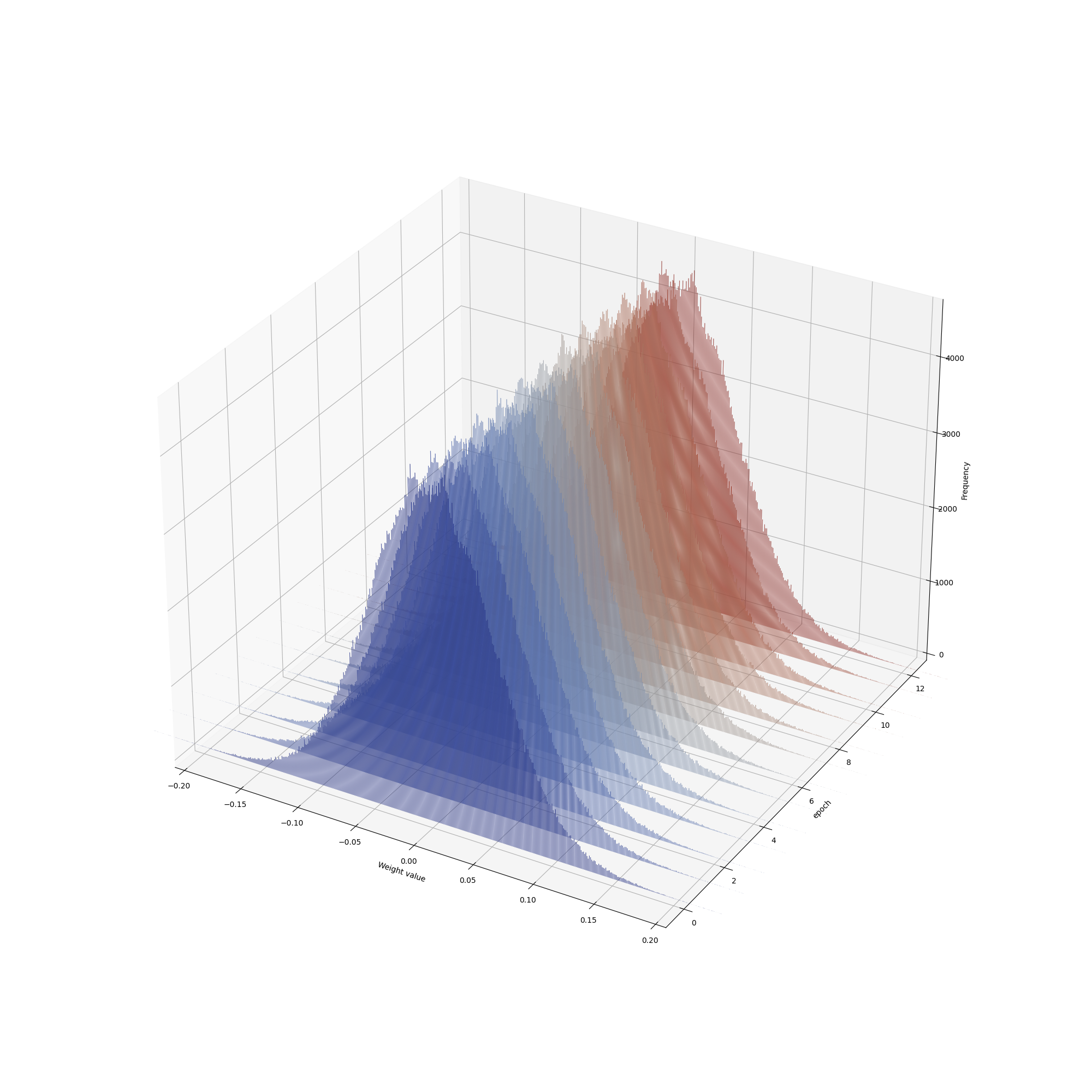}
    \caption{weight distribution of Progressive QAT}
    \label{fig:neck_qat_progressive_ablation}
  \end{subfigure}
  \hfill
  \begin{subfigure}{0.33\linewidth}\centering
    \includegraphics[width=1.2\linewidth]{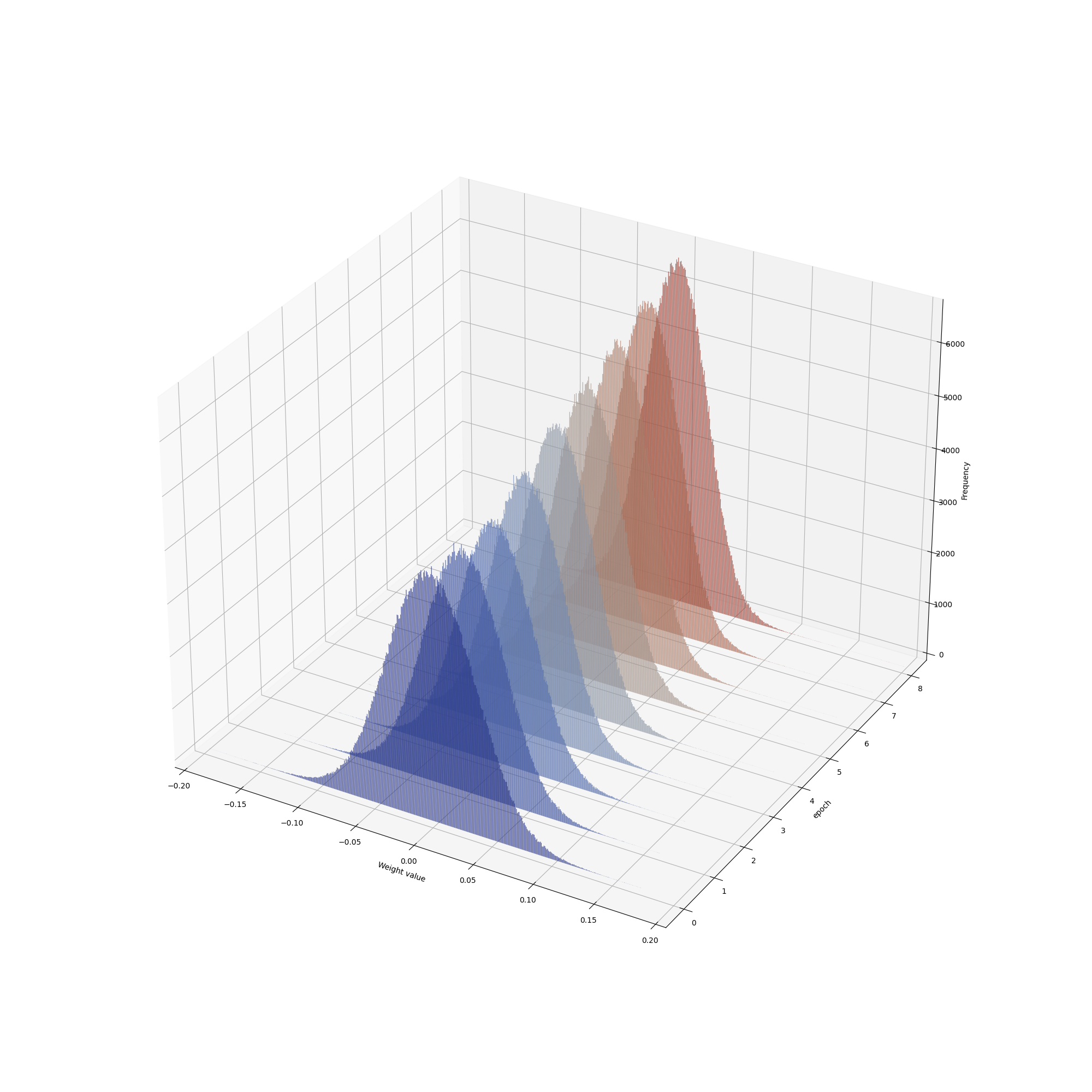}
    \caption{weight distribution of Standard QAT}
    \label{fig:neck_vgd_ablation}
  \end{subfigure}
    \centering
  \caption{Visualization of the weight distribution shift during VGD, progressive QAT, and standard QAT, measured on the image neck. The x-axis stands for the weight value, the y-axis shows the index of the current epoch during training (either VGD, progressive QAT, or standard QAT), and the z-axis is the frequency of the corresponding weight value.}
  \label{fig:neck_weight_distribution}
\end{figure*}

\begin{figure*}[htbp]
  \centering
  \begin{subfigure}{0.33\linewidth}\centering
    \includegraphics[width=1.2\linewidth]{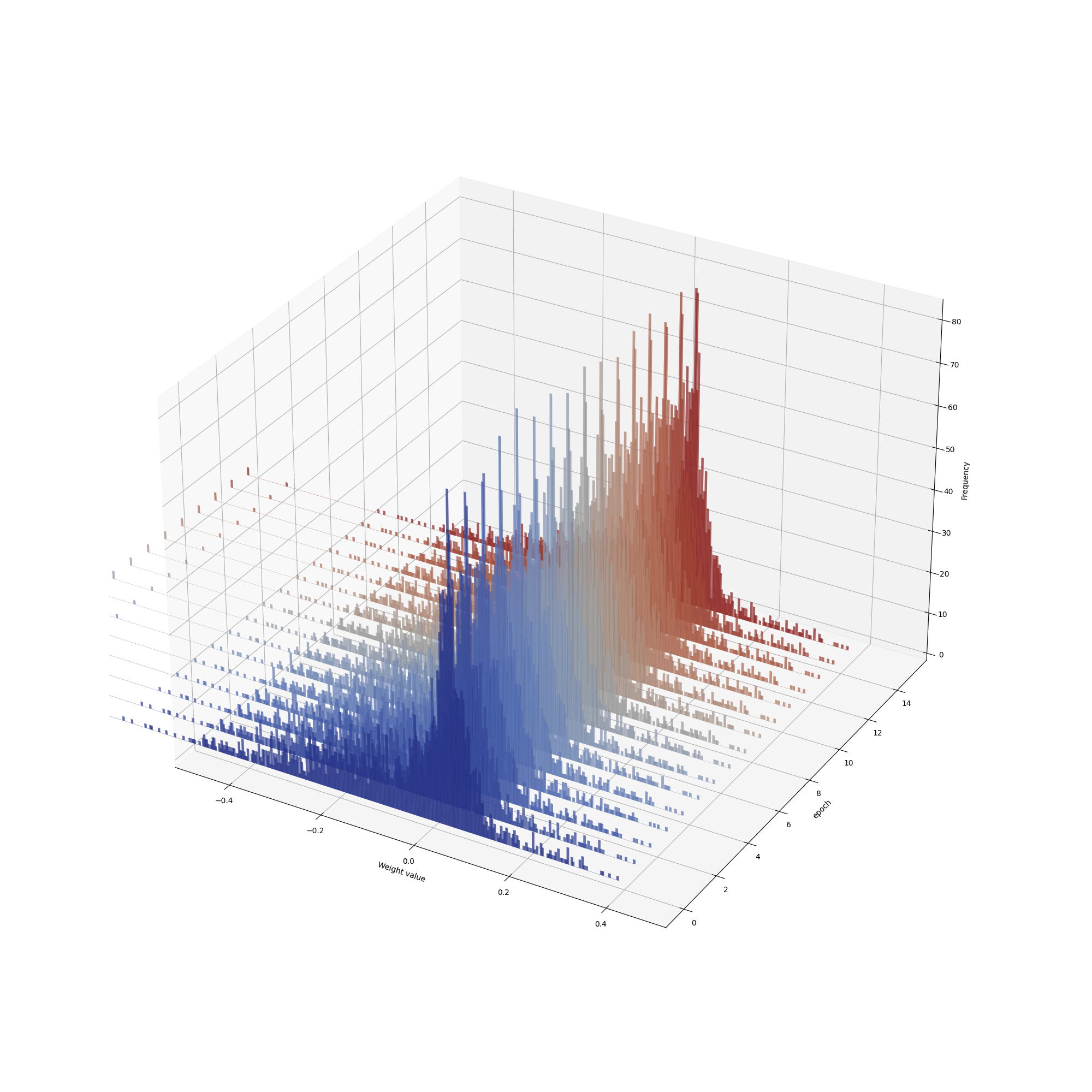}
    \caption{weight distribution of VGD}
    \label{fig:cls_branch_qat_a4_ablation}
  \end{subfigure}
  \hfill
  \begin{subfigure}{0.33\linewidth}\centering
    \includegraphics[width=1.2\linewidth]{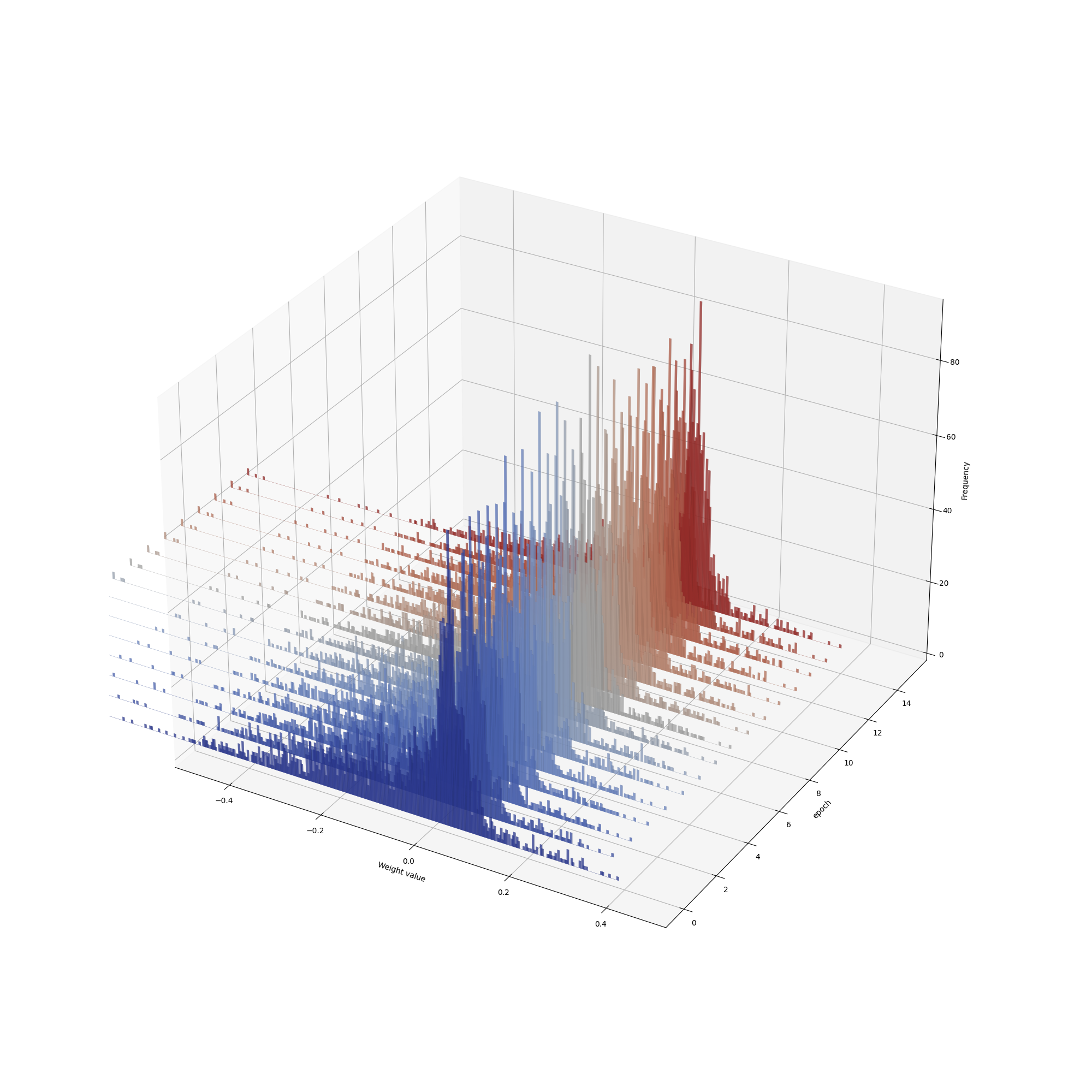}
    \caption{weight distribution of Progressive QAT}
    \label{fig:cls_branch_qat_progressive_ablation}
  \end{subfigure}
  \hfill
  \begin{subfigure}{0.33\linewidth}\centering
    \includegraphics[width=1.2\linewidth]{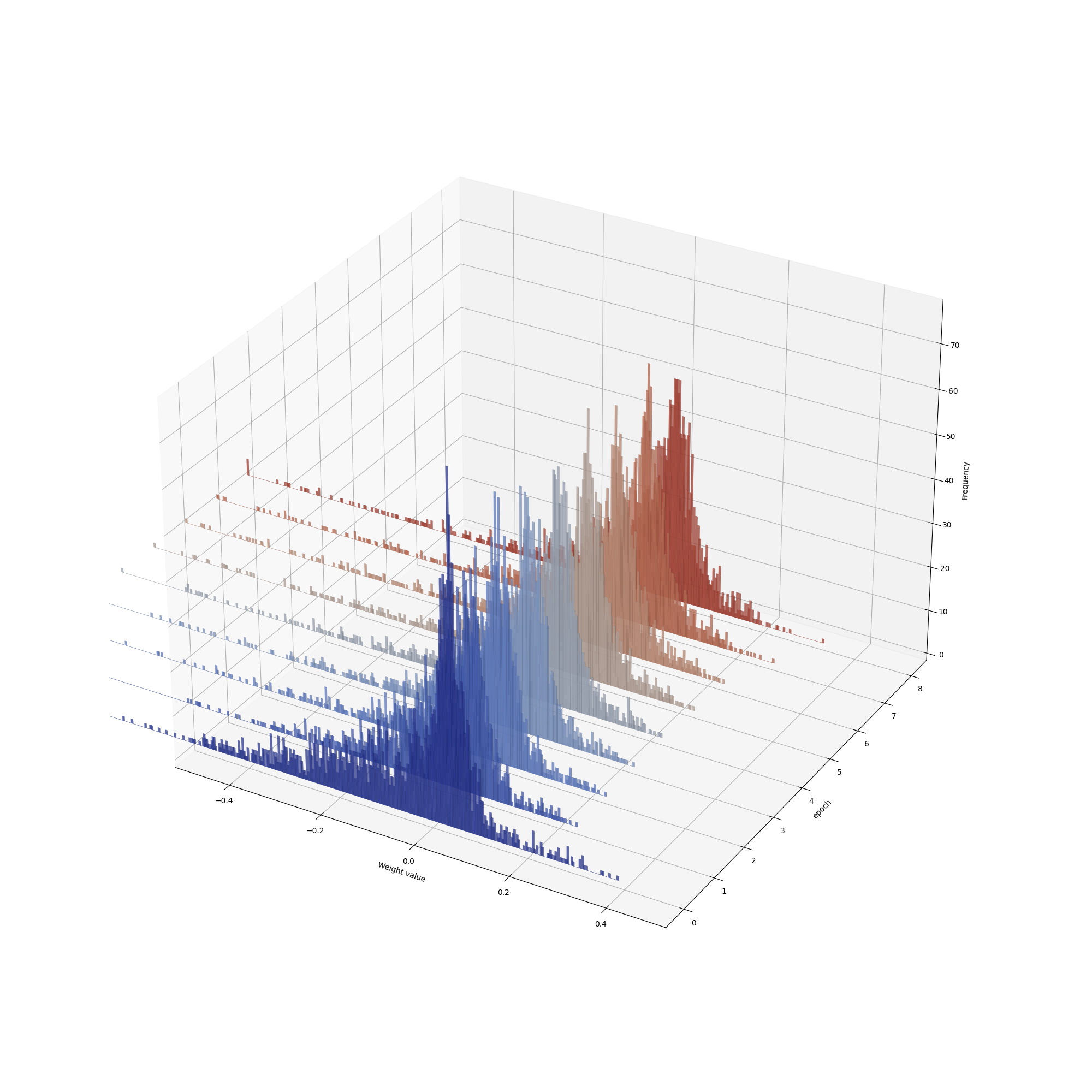}
    \caption{weight distribution of Standard QAT}
    \label{fig:cls_branch_vgd_ablation}
  \end{subfigure}
    \centering
  \caption{Visualization of the weight distribution shift during VGD, progressive QAT, and standard QAT, measured on the classification branch. The x-axis stands for the weight value, the y-axis shows the index of the current epoch during training (either VGD, progressive QAT, or standard QAT), and the z-axis is the frequency of the corresponding weight value.}
  \label{fig:cls_branch_weight_distribution}
\end{figure*}

\section{Additional Visualization}
\subsection{Feature map visualization}

We visualize the feature map of the QD-BEV-Base model with view-guided distillation in Figure~\ref{fig:visusl_feature_map}. Here the teacher is BEVFormer-Base, and the student is our QD-BEV-Base model with 4-bit weights and 6-bit activations. Our view-guided distillation method considers the Image feature and the BEV feature at the same time to obtain a better convergence direction of the model.

\subsection{Visualization on the oscillation during QAT}
As discussed in the main contexts, we found that the quantization-aware training (QAT) has severe stability issues, with accuracy curves of both mAP and NDS oscillating up and down throughout the process. We visualize this phenomenon in Figure~\ref{fig:oscillation}. As shown in Figure~\ref{fig:oscillation of Standard QAT}, when conducting QAT for W4A6, the standard QAT sometimes suffers from gradient explosion, causing the training to collapse after a few epochs. In contrast, the progressive QAT has a better curve, but the stability issue still exists, and it has the drawback that it is difficult to achieve higher accuracy, as shown in Figure~\ref{fig:oscillation of Progressive QAT}. On the other hand, the VGD curve maintains stability while continuously improving accuracy, eventually achieving excellent results, as shown in~\cref{fig:oscillation of VGD}.




To explain this phenomenon, we visualize the changes in weight distribution during training. Figure~\ref{fig:neck_weight_distribution} shows the weight distribution of the image neck during training. As can be seen, for standard QAT in Figure~\ref{fig:neck_qat_a4_ablation}, it collapses since the weights converge to zero during the training. Compared with progressive QAT in Figure~\ref{fig:neck_qat_progressive_ablation}, VGD in Figure~\ref{fig:neck_vgd_ablation} gradually learns weight distributions that can extract more features, while the weight distribution of progressive QAT does not change significantly.

Figure~\ref{fig:cls_branch_qat_progressive_ablation} shows the weight distribution of the classification branch during training. We can see that VGD in Figure~\ref{fig:cls_branch_vgd_ablation} is more stable and does not experience irregular oscillations in weight distribution like progressive QAT in Figure~\ref{fig:cls_branch_qat_progressive_ablation}.

\begin{figure*}[htbp]
  \centering
  \begin{subfigure}{1\linewidth}\centering
    \includegraphics[width=0.7\linewidth]{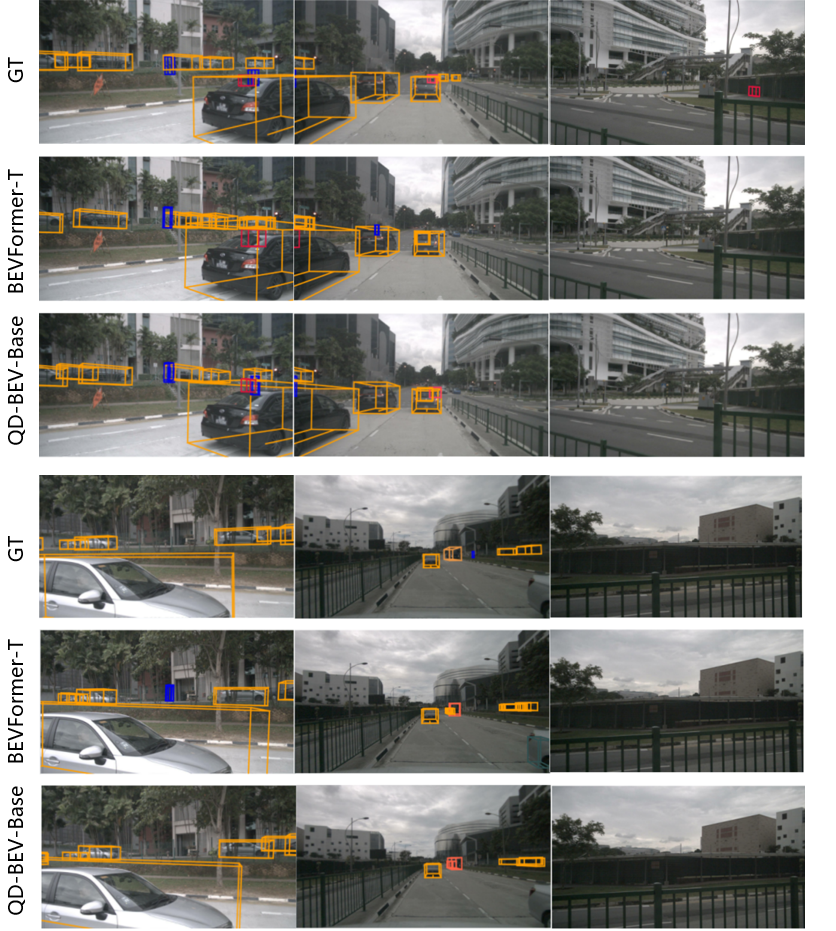}
    \caption{Visualization of 3D detection results of QD-BEV-Base, BEVFormer-T and Ground Truth}
    \label{fig:3d}
  \end{subfigure}
  \hfill
  \begin{subfigure}{1\linewidth}\centering
    \includegraphics[width=0.7\linewidth]{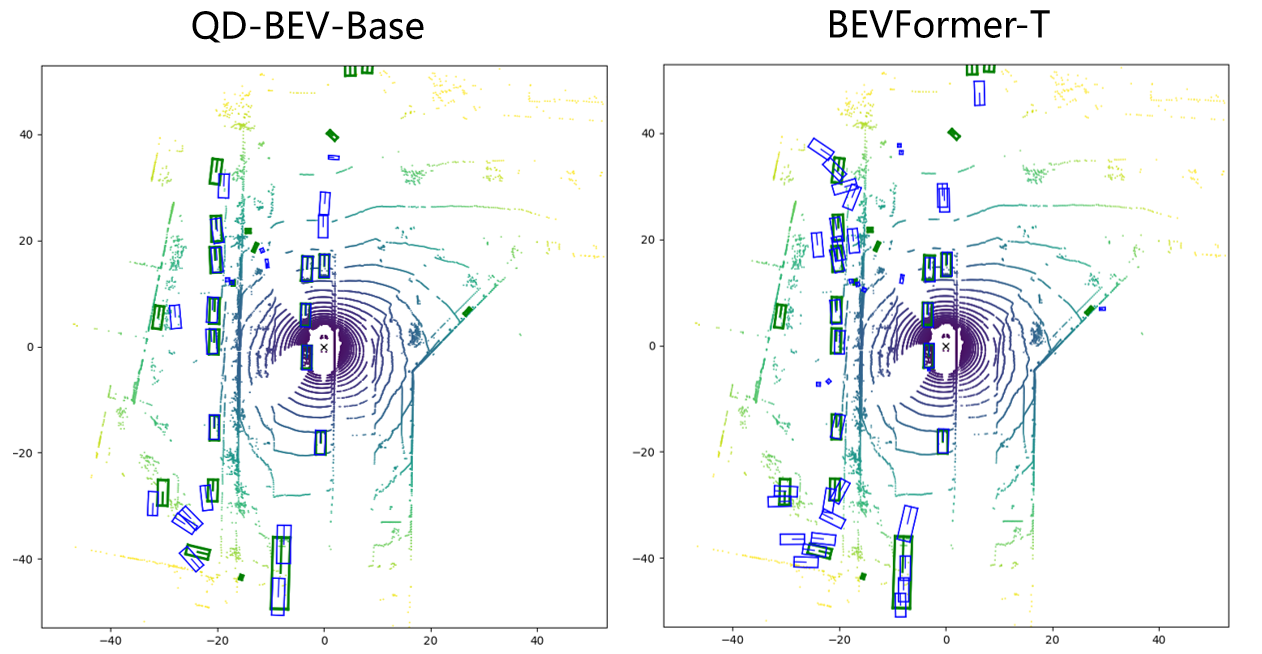}
    \caption{BEV visualization of QD-BEV-Base and BEVFormer-T}
    \label{fig:bev}
  \end{subfigure}
    \centering
  \caption{Visualization of QD-BEV-Base results and the comparison
with results obtained by BEVFormer-T and Ground Truth. The upper figures are from front cameras, and the lower figures are from back cameras.}
  \label{fig:Visual_sec7}
\end{figure*}

\subsection{Additional 3D object detection visualization}

Here in Figure~\ref{fig:Visual_sec7}, we visualize our model on more samples.
In Figure~\ref{fig:3d}, the upper part is the viewing angle of the three cameras in front of the car, and the lower half is the viewing angle of the three cameras behind the car. From top to bottom are the Ground Truth, BEVFormer-Tiny results, and our QD-BEV-Base results, respectively. We can find that QD-BEV-Base predicts the results more accurately than BEVFormer-T, and the error rate is also significantly lower.

In Figure~\ref{fig:bev}, the BEV visualization of the QD-BEV-Base is on the left and the BEVFormer-Tiny result is shown on the right as a comparison. In these two pictures, the color blue represents the predicted results and the color green represents the Ground Truth. As can be seen in the figures, QD-BEV-Base (32.9MB) made a much clearer and more accurate prediction than BEVFormer-Tiny (126.8MB) with only 1/4 of the model size.

\section{Video Demo}

To showcase the efficacy of QD-BEV models, we made a 1080P HD video demo of QD-BEV-Base (32.9 MB) for about 20 seconds, including its comparison with BEVFormer-T (126.8 MB) and Ground Truth. 
The video is attached to the zip file.


{\small
\bibliographystyle{ieee_fullname}
\bibliography{latex/egbib_appendix}
}